\documentclass[preprint,12pt,authoryear]{elsarticle}

\usepackage{amssymb}
\usepackage{amsmath}
\usepackage{pdflscape}
\usepackage{longtable}
\usepackage{tabularx}               
\usepackage{array}    
\usepackage{natbib}

\journal{}
\usepackage[colorlinks=true,linkcolor={0 0 0.5},citecolor={0 0 0.5},urlcolor={0 0 0.5}]{hyperref}

\begin{document}

\begin{frontmatter}



\title{A Comparative Evaluation of Pre-trained Convolutional Neural Networks for Melanoma Detection} 



\author[inst1]{Wagner Moreno Schmitz}
\author[inst1]{Marco Antonio de Castro Barbosa}
\author[inst2]{Thiago Magalhães Amaral}
\author[inst1]{Dalcimar Casanova}
\author[inst1]{Jefferson Tales Oliva}

\affiliation[inst1]{organization={Universidade Tecnológica Federal do Paraná (UTFPR)}, 
            city={Pato Branco},
            state={Paraná},
            country={Brasil}}

\affiliation[inst1]{organization={Universidade Federal do Vale do São Francisco (Univasf)}, 
            city={Petrolina},
            state={Pernambuco},
            country={Brasil}}

\begin{abstract}
Early diagnosis of melanoma is critical for improving patient survival rates. However, accurately distinguishing melanoma from other skin lesions remains a significant clinical challenge due to the high visual similarity among lesion types and variability in image acquisition conditions. Artificial intelligence, particularly machine learning, has emerged as a promising tool to support dermatological diagnosis by automating feature extraction from medical images. Among the available approaches, convolutional neural networks (CNNs) have demonstrated strong performance in image classification tasks, making them well-suited for analyzing both dermatoscopic and histopathological images, given their ability to capture hierarchical visual patterns relevant to lesion characterization. Nevertheless, despite numerous pre-trained CNN architectures having been proposed, selecting the most appropriate one for a given imaging modality remains an open challenge. In this study, we evaluate pre-trained convolutional neural networks (CNNs) for skin lesion classification using dermatoscopic and histopathological image datasets. Experiments were conducted on the HAM10000, ISIC 2018, and CR-AI4SkIN datasets, evaluating the ResNet50, VGG16, VGG19, MobileNet, and InceptionV3 architectures under the same training protocol. The experimental evaluation showed that the models achieved accuracies ranging from 71\% (InceptionV3 on ISIC 2018) to 84\% (ResNet50 on HAM10000) on dermatoscopic images. For histopathological images, accuracies ranged from 72\% (VGG19) to 83\% (ResNet50) on the CR-AI4SkIN dataset. The results demonstrate that model performance differs between dermatoscopic and histopathological image modalities, showing that architectures exhibiting similar performance on dermatoscopic images exhibit different performance on histopathological data.. These findings emphasize the importance of considering image modality when selecting deep learning architectures for melanoma detection.
\end{abstract}


\end{frontmatter}

\section{Introduction}
Skin cancers constitute the most commonly diagnosed group of neoplasms worldwide, representing one of the major current challenges for global health. According to the International Agency for Research on Cancer (IARC), the cancer research arm of the World Health Organization (WHO), the term ``skin cancer'' encompasses a spectrum of diseases that includes both non-melanoma skin carcinomas and malignant melanoma \citep{WHO_skin_cancer}.
Melanoma is recognized as the most aggressive form of skin cancer due to its high metastatic potential and the elevated mortality rates associated with advanced stages of the disease \citep{Pekarek2025}.

Recent epidemiological estimates indicate a concerning global trend of increasing melanoma incidence \citep{Langselius2025}. As reported by \citep{Karimkhani2022}, the number of melanoma cases is expected to increase by more than 50\% between 2020 and 2040. Based on demographic changes and population aging, it is projected that more than 500,000 new cases and approximately 100,000 annual deaths from melanoma will occur worldwide by 2040. Despite this alarming scenario, the same study emphasizes that a substantial proportion of these cases could be prevented through effective public health strategies, including early detection.

Early-stage melanoma detection is decisive for patient survival: approximately 99\% of patients diagnosed at an early stage in the United States survive at least five years after detection \citep{skincancer2025melanoma}. However, the accuracy of naked-eye diagnosis of cutaneous melanoma is only around 60\%. Techniques such as dermoscopy, a non-invasive examination that enables the visualization of skin structures not visible to the naked eye, can substantially increase diagnostic accuracy. Nevertheless, its effectiveness strongly depends on the examiner's level of expertise \citep{kittler2002diagnostic}.

In addition, in some cases, detection can be performed through the analysis of histopathological images obtained from tissue samples removed by biopsy and examined under a microscope. These images provide detailed information about cellular morphology and tissue architecture, enabling the identification of specific characteristics associated with melanoma. According to the guidelines \citep{coelho2025guidelines}, histopathological examination is considered the gold standard for diagnostic confirmation, although it involves manual procedures and subjective interpretation by pathologists.

Advances in technology have contributed to making melanoma detection faster and more accurate. Methods based on artificial intelligence have shown promising results, assisting in screening and diagnostic support \citep{PARHI2026193}. Nevertheless, challenges persist related to data variability, the heterogeneous performance of neural network architectures, and model generalization capability. In this context, the evaluation of different CNN architectures under controlled experimental conditions becomes essential to identify models that are more robust and reliable for supporting medical diagnosis. Current studies commonly evaluate models on a single dataset or imaging modality, or employ different training protocols and hyperparameter configurations, making it difficult to isolate the effects of the CNN architecture from those of the experimental setup. Therefore, this study evaluates the performance of pre-trained CNN architectures using an identical training protocol and hyperparameter configuration across dermatoscopic and histopathological image datasets, enabling a fair assessment of how imaging modality influences the performance and generalization capability of different CNN architectures for automatic melanoma detection.

Among these technologies, \cite{athiwaratkun2015feature} point out that convolutional neural networks (CNN) stand out due to their ability to automatically extract relevant patterns and characteristics from images. Since their popularization, numerous architectures have been proposed, such as VGG, ResNet, Inception, and DenseNet, each with distinct designs that influence their capacity to learn discriminative representations. However, training these models from scratch typically requires large annotated datasets and substantial computational resources, which are often scarce in the medical imaging domain. To address this limitation, pre-trained networks have been widely adopted through transfer learning, a strategy in which models previously trained on large-scale datasets, such as ImageNet, are fine-tuned for a specific target task with considerably less data.

In this way, the analysis of dermoscopic and histopathological images using CNNs overcomes the limitations of manual evaluation, which can be subjective and dependent on examiner experience. These networks are capable of identifying subtle details that often go unnoticed in conventional analysis. Comparing different CNN architectures allows for the assessment of how each model responds to data variability, the identification of more robust and generalizable models, and the selection of those with greater potential to improve automatic melanoma detection, contributing to more accurate and consistent diagnoses.

In this work, we evaluate five pre-trained CNN architectures for melanoma detection across three distinct datasets: two composed of dermatoscopic images and one of histopathological images. The inclusion of two dermatoscopic datasets enables the assessment of model consistency across different datasets within the same imaging modality, while the histopathological dataset allows the evaluation of model behavior under a distinct imaging modality. The adopted strategy aims to evaluate the behavior of different architectures on both homogeneous and heterogeneous datasets, investigating whether there are statistically significant differences among models and how the imaging modality influences the obtained results. This analysis enables the identification of architectures that exhibit greater robustness, generalization capability, and consistent performance across different image types, providing important insights into model selection for clinical applications of automatic melanoma detection and contributing to the development of more accurate and reliable medical decision-support systems.

The main contributions of this work are as follows:

\begin{itemize}
\item A systematic comparison of five CNN architectures employed as feature extractors across different image datasets, considering both dermoscopic and histopathological data;

\item The evaluation of performance and generalization capability across modalities and multiple datasets enables an understanding of how each architecture responds to the structural variations present in distinct data sources;

\item The application of statistical hypothesis testing, specifically the Friedman and Nemenyi tests, to rigorously assess whether performance differences among architectures are statistically significant;

\end{itemize}

Subsequently, the remainder of this paper is organized as follows: Section \ref{sec:Related-Work} presents the Related Work, including a systematic review of related studies and the identification of methodological patterns and limitations in the literature; Section \ref{sec:Materials-and-Methods} describes the materials and methods adopted in this study, detailing the datasets used (HAM10000, ISIC 2018 Challenge, and CR-AI4SkIN), the preprocessing procedures, the evaluated pre-trained convolutional neural network architectures, the feature embedding extraction process, and the evaluation metrics; Section \ref{sec:Results-and-Discussion} reports and discusses the experimental results, analyzing the performance of the evaluated architectures across different datasets and including statistical comparisons based on the Friedman and Nemenyi tests; and Section \ref{sec:Conclusion} concludes the paper by summarizing the main findings, discussing their implications, and outlining directions for future research in automatic melanoma detection.

\section{Related Work}
\label{sec:Related-Work}

The development of this research is grounded in the analysis of recent studies that employ machine learning and deep learning techniques for skin cancer detection. Based on this review, the objective is to identify the most frequently adopted methodological practices in the literature, as well as to recognize recurring patterns and limitations that may be addressed or overcome. This stage is fundamental for defining the parameters and guidelines to be adopted in this study, with the goal of implementing a robust model with high generalization potential for automated diagnosis.

To ensure a systematic and reproducible selection of the literature, the bibliographic review followed specific criteria. The search was conducted in the IEEE Xplore database, chosen for its relevance in computer science and technology, and complemented by a search in the journal Expert Systems with Applications, selected for its relevance in applied computational research, despite not offering open-access publications. The search string was constructed using logical combinations of keywords in English: ("CNN" OR "Convolutional Neural Network") AND ("Image Classification" OR "Medical Images") AND ("Deep Learning") AND ("skin lesions" OR "skin cancer").

After retrieving the initial search results, a sequential screening protocol was applied to determine the eligibility of each study. The selection process consisted of five verification steps: (i) the study had to be related to the healthcare domain; (ii) it had to be written in English; (iii) it had to employ a Convolutional Neural Network (CNN) as the primary methodological approach; (iv) for the IEEE Xplore search, it had to be available with full open access, this criterion was not applied to the search conducted in Expert Systems with Applications, given that the journal does not provide open-access articles; and (v) it had to address skin cancer or related cancer classification problems. Only studies that satisfied all applicable criteria were included in the detailed analysis.

For the refinement of the results, the following inclusion criteria were applied: (i) articles written in English; (ii) studies addressing the application of CNN for the classification of skin cancer medical images; (iii) publications between January 2023 and May 2026; and (iv) articles available in full text. Conversely, studies that did not use CNNs as the primary method, articles without full-text access, and those with titles or abstracts unrelated to the research theme were excluded.

Following the initial search, a total of 278 articles were identified. After the application of the selection protocol and inclusion criteria, 27 articles were selected for detailed analysis. These studies provide a comparative baseline for the current research, focusing on state-of-the-art implementations within the specified period.

An overview of the analyzed articles, based on the previously established criteria: Dataset, Image Set Size, Problem Type, Training Testing Strategy, Applied Model, Main Features, Evaluation Metrics, Reported Results, and Frameworks, is provided in Table~\ref{tab:revisao_literatura}. To ensure a clear understanding of the comparative data, the definitions and parameters considered for each column of the table are described below:

\begin{itemize}
    \item Work: Authors and year of publication.
    \item Image set: image dataset used in each study.
    \item Feature categories: Category or type of features employed.
    \item Classification algorithms: Algorithms or neural network architectures used for classification.
    \item Performance measures: Reported performance evaluation measures used in the related work.
    \item Fine-tuning / Transfer learning: Indication of whether fine-tuning or transfer learning was applied, and which pre-trained models were used.
    
\end{itemize}

\begingroup
\tiny
\renewcommand{\arraystretch}{1.5}
\begin{longtable}{|p{1.5cm}|c|c|c|c|c|c|c|c|}

\caption{Systematic literature review of skin cancer classification methods.} \label{tab:revisao_literatura} \\
\hline
\begin{tabular}[c]{@{}c@{}}\textbf{Work} \end{tabular} & \begin{tabular}[c]{@{}c@{}} \textbf{Image} \textbf{set} \end{tabular} & \begin{tabular}[c]{@{}c@{}} \textbf{Feature}\\ \textbf{categories}\end{tabular} & \begin{tabular}[c]{@{}c@{}} \textbf{Classification}\\\textbf{algorithms} \end{tabular} & \begin{tabular}[c]{@{}c@{}}\textbf{Performance}\\  \textbf{measures}\end{tabular} & \begin{tabular}[c]{@{}c@{}}\textbf{Fine-tuning/}\\ \textbf{Transfer}\\ \textbf{learning} \end{tabular}\\
\hline
\endfirsthead

\multicolumn{9}{c}{{\tablename\ \thetable{} -- continued from previous page}} \\
\hline
\begin{tabular}[c]{@{}c@{}}\textbf{Work} \end{tabular} & \begin{tabular}[c]{@{}c@{}} \textbf{Image} \textbf{set} \end{tabular} & \begin{tabular}[c]{@{}c@{}} \textbf{Feature}\\ \textbf{categories}\end{tabular} &  \begin{tabular}[c]{@{}c@{}} \textbf{Classification}\\\textbf{algorithms} \end{tabular} & \begin{tabular}[c]{@{}c@{}}\textbf{Performance}\\  \textbf{measures}\end{tabular} & \begin{tabular}[c]{@{}c@{}}\textbf{Fine-tuning/}\\ \textbf{Transfer}\\ \textbf{learning} \end{tabular}\\
\hline
\endhead

\hline
\multicolumn{9}{r}{{Continued on next page}} \\
\endfoot

\hline
\endlastfoot

\cite{10288439} & \begin{tabular}[c]{@{}c@{}}ISIC 2017\\ and 2020\end{tabular} & \begin{tabular}[c]{@{}c@{}}Deep and\\ Metaheuristic-based\\ Features\end{tabular} & HMDL-MFMBIA & \begin{tabular}[c]{@{}c@{}}Accuracy,\\ F1-score,\\ precision,\\ recall, and\\ AUC\end{tabular} & Yes \\
\hline
\cite{10374026} & \begin{tabular}[c]{@{}c@{}}HAM10000\\NCT-CRC-HE-100K\end{tabular} & \begin{tabular}[c]{@{}c@{}}Attention, and\\ multi-scale features\end{tabular} & SPCB-Net & \begin{tabular}[c]{@{}c@{}}Accuracy (Acc),\\ and F1-score (F1)\end{tabular} & Yes \\
\hline
\cite{10416885} & HAM10000 & GAN and attention  & \begin{tabular}[c]{@{}c@{}}STGAN and\\ T-ResNet50\end{tabular} & \begin{tabular}[c]{@{}c@{}}Sensitivity (Sen),\\ Acc, F1, and\\ specificity (Spe)\end{tabular} & Yes \\
\hline

\cite{Wang2022} & \begin{tabular}[c]{@{}c@{}}ISIC 2016, \\2017, and\\ 2018, and\\ PH2\end{tabular} & \begin{tabular}[c]{@{}c@{}}Cascaded\\ Context\\ Aggregation (CCA),\\ and Context-Guided\\ Local affinity (CGL)\end{tabular} & \begin{tabular}[c]{@{}c@{}}ResNet-ASPP,\\ CCA, and CGL\end{tabular} & \begin{tabular}[c]{@{}c@{}}Jaccard (Jac)\\ and dice\end{tabular} & Yes \\\hline

\cite{10416953} & \begin{tabular}[c]{@{}c@{}}ISIC 2017, \\2019, and\\ 2020\end{tabular} & \begin{tabular}[c]{@{}c@{}}Shape and Texture\end{tabular} & \begin{tabular}[c]{@{}c@{}} k-means, NCC,\\ RF, and SVM\end{tabular} & \begin{tabular}[c]{@{}c@{}}Acc, F1,\\ precision (Pre),\\ and recall (Rec)\end{tabular} & No \\
\hline
\cite{10695064} & \begin{tabular}[c]{@{}c@{}}ISIC 2019\\ HAM10000\end{tabular} & \begin{tabular}[c]{@{}c@{}}Fusion features\end{tabular} & EFAM-Net & \begin{tabular}[c]{@{}c@{}}Acc and F1\end{tabular} & Yes \\
\hline
\cite{10965692} & \begin{tabular}[c]{@{}c@{}}ISIC 2017\\ HAM10000\end{tabular} & \begin{tabular}[c]{@{}c@{}}Global/Local Attention,\\ and Multi-scale Fusion\end{tabular} & \begin{tabular}[c]{@{}c@{}}EG-VAN\end{tabular} & \begin{tabular}[c]{@{}c@{}}Acc, Rec, and\\ F1\end{tabular} & Yes \\
\hline
\cite{11029577} & \begin{tabular}[c]{@{}c@{}}ISIC 2017 HAM10000\end{tabular} & \begin{tabular}[c]{@{}c@{}}CNN features\end{tabular} & \begin{tabular}[c]{@{}c@{}}DRL-driven\\ Active\\ Learning\end{tabular} & F1 & Yes \\
\hline
\cite{11098782} & \begin{tabular}[c]{@{}c@{}}ISIC 2019 and\\ PAD-UFES-20\end{tabular} & \begin{tabular}[c]{@{}c@{}}Compressed deep\\ features\end{tabular} & \begin{tabular}[c]{@{}c@{}}Optimized\\ AlexNet\end{tabular} & \begin{tabular}[c]{@{}c@{}}Acc, Sen, Spe,\\ Pre, and F1\end{tabular} &  Yes \\
\hline
\cite{11121145} & HAM10000 & \begin{tabular}[c]{@{}c@{}}Deep Residual\\ Features\end{tabular} & \begin{tabular}[c]{@{}c@{}} ResNet-50\end{tabular} & \begin{tabular}[c]{@{}c@{}}Acc, Pre, Rec,\\ and F1\end{tabular} & Yes \\
\hline
\cite{11121108} & \begin{tabular}[c]{@{}c@{}}ISIC 2016, \\2017, and \\2018\end{tabular} & \begin{tabular}[c]{@{}c@{}}Multi-Scale Spatial\\ and Contextual\end{tabular} & UniSegNet & Dice and IoU & Yes \\
\hline
\cite{11129704} & \begin{tabular}[c]{@{}c@{}}ISIC 2016,\\ 2017, and\\ 2018\end{tabular} & \begin{tabular}[c]{@{}c@{}}Multi-model deep\\ features\end{tabular} & \begin{tabular}[c]{@{}c@{}}XceptMPX\end{tabular} & \begin{tabular}[c]{@{}c@{}}Acc, Pre, Rec,\\ F1, and area\\ under curve (AUC)\end{tabular} & Yes \\
\hline


\cite{11263785} & \begin{tabular}[c]{@{}c@{}}ISIC 2017\\and PH2\end{tabular} & \begin{tabular}[c]{@{}c@{}}Multi-task features\end{tabular} & CA Y-Net &\begin{tabular}[c]{@{}c@{}} Acc, Sen, Spe,\\ and AUC\end{tabular} & Yes \\
\hline
\cite{11284879} & \begin{tabular}[c]{@{}c@{}}HAM10000\\ ISIC 2019\\and\\ BCN20000\end{tabular} & \begin{tabular}[c]{@{}c@{}}Ensemble-based deep\\ features\end{tabular}  & SynthraXCoreNet & \begin{tabular}[c]{@{}c@{}}Acc, Rec, Pre,\\ and F1\end{tabular} & Yes \\
\hline
\cite{11354481} & \begin{tabular}[c]{@{}c@{}}ISIC 2016,\\ 2017, and\\ 2018, and\\ PH2\end{tabular} & \begin{tabular}[c]{@{}c@{}}Global, local, and\\ region-level cues\end{tabular} & GLR-Net  & \begin{tabular}[c]{@{}c@{}}Dice, Acc, and\\ intersection\\ over union (IoU)\end{tabular} & Yes \\
\hline
\cite{Alenezi2023} & \begin{tabular}[c]{@{}c@{}} ISIC 2019 \\and 2020\end{tabular} & \begin{tabular}[c]{@{}c@{}} Deep Feature\\ Extraction\end{tabular} & \begin{tabular}[c]{@{}c@{}}Bayesian-\\optimized\\ SVM \end{tabular} & \begin{tabular}[c]{@{}c@{}}Acc, Sen, Spe,\\ and Pre\end{tabular} & Yes \\
\hline
\cite{Hu2022} & \begin{tabular}[c]{@{}c@{}}ISIC 2017\\ and 2018,\\and PH2\end{tabular} & \begin{tabular}[c]{@{}c@{}}Spatial and Channel\\ Attention\end{tabular} & AS-Net & Acc, and Jac & Yes \\
\hline

\cite{Abdelhalim2021} & HAM10000 & \begin{tabular}[c]{@{}c@{}}GAN-based\\ features\end{tabular} & ResNet-18 & \begin{tabular}[c]{@{}c@{}}Acc, Sen, Rec,\\ Pre, and F1\end{tabular} & Yes\\
\hline
\cite{Zhu2025} & \begin{tabular}[c]{@{}c@{}}ISIC 2016,\\ 2017, and\\ 2018, and\\ PH2 PAD-\\UFES-20\end{tabular} & \begin{tabular}[c]{@{}c@{}}Morphology-aware \\boundary extraction,\\ multi-scale fusion,\\ and few-shot domain\\ generalization\end{tabular} & EM-Net & \begin{tabular}[c]{@{}c@{}}Acc, Pre, Rec,\\ Dice and IoU\end{tabular} & Yes \\
\hline

\cite{Kaymak2020} & ISIC 2017 & \begin{tabular}[c]{@{}c@{}}FCN-based\\ features\end{tabular} & FCN-8s & \begin{tabular}[c]{@{}c@{}}Acc, Dice,\\ and Jac\end{tabular} & Yes \\
\hline

\cite{Reis2026} & PH2 & \begin{tabular}[c]{@{}c@{}}Multi-scale attention,\\and ensemble fusion\end{tabular} & \begin{tabular}[c]{@{}c@{}}SVM, RF, and\\ KNN hard-\\voting\\ ensemble\end{tabular} & \begin{tabular}[c]{@{}c@{}}Acc, Pre, Sen,\\ and F1\end{tabular} & Yes\\
\hline
\cite{Zhang2026} & \begin{tabular}[c]{@{}c@{}}ISIC 2018\\ and 2019 \end{tabular}& \begin{tabular}[c]{@{}c@{}}Multi-scale attention\\ fusion\end{tabular} & GL-MSFN & \begin{tabular}[c]{@{}c@{}}Acc, Pre, Rec,\\ and Spe\end{tabular} & Yes\\
\hline
\cite{Aruk2025} & \begin{tabular}[c]{@{}c@{}}HAM10000\\ and\\ ISIC 2019\end{tabular} &\begin{tabular}[c]{@{}c@{}}Local-global \\ hybrid (CNN-ViT)\end{tabular} & Hybrid CNN-ViT & Acc and F1 & Yes\\
\hline
\cite{PezhmanPour2020} & \begin{tabular}[c]{@{}c@{}}ISIC 2016\\ and 2017\end{tabular} & \begin{tabular}[c]{@{}c@{}}Contourlet-transform,\\ CIELAB color\end{tabular} & \begin{tabular}[c]{@{}c@{}}Compact encoder-\\decoder CNN\end{tabular} & \begin{tabular}[c]{@{}c@{}}Jac, Sen, and\\ dice\end{tabular} & No  \\
\hline
\cite{Sethanan2023} & \begin{tabular}[c]{@{}c@{}} HAM10000\\ and HAM-MB-\\13312\end{tabular} & \begin{tabular}[c]{@{}c@{}}Ensemble deep features\end{tabular} & \begin{tabular}[c]{@{}c@{}}AMIS-weighted\\ double-ensemble\end{tabular} & \begin{tabular}[c]{@{}c@{}}Acc, AUC, and\\ Pre\end{tabular} & Yes \\
\hline
\cite{Sulthana2024} & HAM10000 & \begin{tabular}[c]{@{}c@{}}Fractals\end{tabular} & S-MobileNet & \begin{tabular}[c]{@{}c@{}}Acc, Pre, and\\ F1\end{tabular} & Yes \\
\hline
\cite{Gao2026} & \begin{tabular}[c]{@{}c@{}}ISIC 2017\\ and 2018,\\ and PH2\end{tabular} & \begin{tabular}[c]{@{}c@{}}Multi-scale\\ transformer-CNN dual\\ encoder\end{tabular} & AMST-Net & \begin{tabular}[c]{@{}c@{}}Dice, IoU, and Acc\end{tabular} & No \\
\hline

\end{longtable}
\endgroup

The analysis of the selected studies presented in Table~\ref{tab:revisao_literatura} revealed a recurring pattern in dataset selection, with HAM10000 being the most frequently used dataset, followed by different editions of the ISIC dataset (2017, 2018, 2019, and 2020). In addition to these public datasets, some studies employed private datasets, which are often not publicly available, thereby limiting reproducibility and hindering direct comparison among methods.

The data splitting strategy was predominantly performed using the hold-out method, adopting proportions such as 80\% for training and 20\% for testing, or alternatively 80\% for training, 10\% for validation, and 10\% for testing.

\cite{10288439} propose the HMDL-MFMBIA framework, a hybrid architecture that combines Swin-UNet-based segmentation, fusion-based feature extraction using Xception and ResNet18, metaheuristic hyperparameter optimization via a Hybrid Salp Swarm Algorithm, and GRU-based classification for biomedical image analysis. The study is evaluated on the ISIC 2017 and ISIC 2020 datasets, each containing 3,000 images across two classes (benign and melanoma), using a 70/30 hold-out split, with performance measured through Accuracy, Precision, Recall, F-Score, and AUC. The proposed method achieves 94.51\% and 95.38\% accuracy on ISIC 2017 and ISIC 2020 respectively, outperforming baselines such as ResNet18, InceptionV3, AlexNet, SVM, and Ensemble models. From a methodological standpoint, the work illustrates how the integration of automated hyperparameter tuning and multi-model feature fusion can yield more robust classification pipelines, though the binary class setup and relatively small dataset size represent limitations worth considering when contextualizing its contribution within broader skin cancer detection research.

Addressing the persistent challenge of intra-class variability and inter-class similarity in dermatological images, \cite{10374026} introduces SPCB-Net (Self-Interactive Attention Pyramid and Cross-Layer Bilinear-Trilinear Pooling Network), a multi-scale architecture combining a Self-Interactive Attention Pyramid, cross-layer bilinear-trilinear pooling, and a Global Average Algorithm over ResNet101 and VGG19 backbones. Evaluated on HAM10000 (10,015 images, 7 classes) and NCT-CRC-HE-100K (100,000 patches, 9 classes) through an 80/10/10 split using Accuracy, Precision, Recall, F1-Score, and AUC under PyTorch, the model reaches 97.10\% and 99.87\% accuracy respectively, outperforming the prior state-of-the-art by 0.4\% on both datasets, demonstrating that high-order feature interaction combined with multi-scale attention yields more discriminative representations for fine-grained medical image classification.

Taking a generative approach to the class imbalance problem, \cite{10416885} proposes STGAN, a two-stage framework that decouples the learning of universal and class-specific knowledge to overcome mode collapse in minority classes, producing diverse 256×256 dermoscopy images for the HAM10000 dataset. The classification pipeline combines the synthetic data with a pretrained ResNet50 and Test-Time Augmentation, evaluated on an 80/20 hold-out split through Accuracy, Sensitivity, Precision, F1-Score, and Specificity under PyTorch, reaching 98.23\% accuracy and 88.85\% sensitivity. Beyond classification, STGAN outperforms conditional StyleGAN2 by 16\% in FID and 33\% in Recall, suggesting that separating global from class-specific generation is a more principled strategy for handling severely imbalanced medical image datasets than conventional conditional GAN approaches.

Unlike the deep learning-centered approaches predominant in the literature, \cite{10416953} takes a classical machine learning route, proposing a normalized cross-correlation-based k-means clustering pipeline for melanoma segmentation and binary classification. The method uses template matching to dynamically determine the number of clusters fed into k-means, followed by GLCM-based Haralick texture \citep{4309314} and Hu-Moment \citep{1057692} shape feature extraction, PCA-based dimensionality reduction, and evaluation across six classifiers including Random Forest, SVM, and KNN. Three ISIC datasets are used (2017, 2019, and 2020) with a 70/30 hold-out split, reporting Accuracy, Precision, Recall, and F1-Score, with the best results reaching 99.46\%, 99.38\%, and 99.29\% respectively across datasets. Notably, the study explicitly benchmarks its results against deep learning models and claims superior performance, which positions it as a relevant reference for discussions on the trade-offs between computational complexity and classification effectiveness in dermoscopic image analysis.

\cite{10608138} propose the IncepX-Ensemble model, a transfer learning-based ensemble combining InceptionV3 and Xception through weighted average fusion for multiclass skin lesion classification. Evaluated on the HAM10000 dataset (10,015 images, 7 classes) using an 80/20 hold-out split and traditional data augmentation to address class imbalance, the model achieves 98.8\% accuracy, outperforming individual baselines such as ResNet50 and InceptionV3, with performance assessed through Accuracy, Precision, Recall, and F1-Score.

\cite{10695064} introduce EFAM-Net, a ConvNeXt-based architecture that incorporates three newly designed modules: an Attention Residual Learning ConvNeXt (ARLC) block for low-level feature extraction, a Parallel ConvNeXt (PCNXt) block for enhanced deep semantic representation, and a Multi-scale Efficient Attention Feature Fusion (MEAFF) block for integrating multi-scale features across network layers. The model is evaluated on ISIC 2019 (25,331 images, 8 classes), HAM10000 (10,015 images, 7 classes), and a private clinical dataset (2,900 images, 6 classes), using an 80/20 hold-out split, with performance assessed through Accuracy, Precision, Recall, Specificity, and F1-Score under PyTorch. Data augmentation techniques, including random rotation, horizontal flipping, and histogram equalization, were applied to the training set to improve robustness. EFAM-Net achieves overall accuracies of 92.30\%, 93.95\%, and 94.31\% on the three datasets, respectively, outperforming baselines such as ResNet-101, DenseNet-201, EfficientNet-B0, and ConvNeXt, while also ranking first across all evaluation metrics on HAM10000 compared to other state-of-the-art methods.

\cite{10734113} address a binary classification problem using the ISIC Archive dataset with only 3,297 dermoscopy images, a considerably small image set compared to other works in the literature, adopting the hold-out strategy with an 80/20 split and applying transfer learning with ImageNet weights across three CNN backbones, namely EfficientNetV2B3, InceptionV3, and InceptionResNetV2, enhanced by a soft attention mechanism. Evaluation was conducted through accuracy, precision, recall, and F1-score, with the best result reaching 95.6\% accuracy. However, the reduced dataset size and binary problem scope limit conclusions about the model's generalization capacity to more complex multi-class clinical scenarios.

Advancing beyond conventional single-strategy approaches, \cite{10965692} propose the EG-VAN architecture, a dual-branch ensemble combining EfficientNetV2S with a modified ResNet50 enhanced by a Spatial-Context Group Attention (SCGA) module and a Non-Local Block, applied to a custom 9-class dataset aggregated from HAM10000 and a binary ISIC dataset totaling over 13,000 images, with a split strategy separating 1,332 original images for testing and the remainder divided into training and validation sets after augmentation. The preprocessing pipeline stands out for its novelty, incorporating hair removal, oversampling, and a color balancing method combining the Gray World algorithm with Retinex theory to enhance lesion visibility. Transfer learning with ImageNet weights was employed across both branches, and focal loss was adopted instead of categorical cross-entropy to address class imbalance. Performance was evaluated through accuracy, precision, recall, F1-score, and AUC, with the model achieving 98.20\% accuracy and 96.68\% F1-score on the 9-class dataset, though generalization to external datasets remains unaddressed, representing a notable limitation for clinical deployment.

Proposing active learning as a solution to the labeled data bottleneck in CNN-based skin cancer detection, \cite{11029577} integrate deep reinforcement learning for dynamic sample selection with a novel scope loss function, evaluated on ISIC (318 images, seven classes) and HAM10000 (10,015 images, seven classes) using a hold-out strategy with preprocessing including resizing, augmentation, and contrast adjustment. An enhanced artificial bee colony algorithm handles hyperparameter optimization, and performance is assessed through accuracy, F-measure, G-means, and AUC, achieving F-measures of 92.791\% and 91.984\% respectively, though the very limited ISIC image set and absence of external validation raise concerns about generalization.

Addressing the challenge of deploying deep learning models on resource-constrained edge devices, \cite{11098782} apply magnitude-based unstructured pruning to a fine-tuned AlexNet for binary skin cancer classification using a small subset of the PAD-UFES-20 dataset (296 images, augmented to 592, split 80:20), focusing on melanoma versus benign nevus distinction. Rather than reporting standard classification metrics as primary results, the study conducts a layer-wise pruning sensitivity analysis, revealing that fully connected layers tolerate aggressive pruning for memory reduction while convolutional layers are more sensitive but dominate computational cost, with 91\% pruning achieving 97.48\% accuracy, 95.83\% sensitivity, and 93.88\% F1-score at a 91\% reduction in parameters, representing a viable trade-off for edge deployment, though the very limited dataset size and absence of comparison against other compressed architectures limit broader conclusions about generalizability.

\cite{11121145}, propose a multistage learning pipeline using ResNet-50 as a static feature extractor for multiclass skin lesion classification on the HAM10000 dataset (10,015 dermoscopic images across seven categories), applying transfer learning with ImageNet weights, targeted data augmentation without SMOTE, and an 80/10/10 hold-out split with 5-fold patient-wise cross-validation. The model achieved 85\% accuracy with macro-averaged precision of 0.75, recall of 0.65, and F1-score of 0.65, evaluated using accuracy, precision, recall, F1-score, and ROC-AUC metrics via PyTorch. The study aligns with the dominant methodological patterns found in the literature, including the use of HAM10000, hold-out splitting, transfer learning, and augmentation-based class balancing, while its modest macro recall reflects the persistent challenge of minority class recognition in imbalanced medical imaging datasets.

\cite{11121108} present UniSegNet, a skin lesion segmentation architecture designed to improve both segmentation accuracy and computational efficiency through the integration of multi-attention mechanisms, asymmetric multi-cross convolutions, and boundary-guided modules. The proposed model was evaluated on the ISIC 2016, 2017, and 2018 datasets using Dice Score, Jaccard Index, pixel accuracy, and inference time as evaluation metrics, achieving Dice scores of 0.935, 0.934, and 0.936, respectively, while requiring only 0.08 seconds per image during inference. Training was conducted in PyTorch using the Adam optimizer with a learning rate of 0.001 over 200 epochs, with images resized to 192×256 and augmented through rotation, flipping, and scaling operations. Although the work targets segmentation instead of classification, it shares several methodological characteristics commonly found in skin lesion analysis studies, such as the adoption of ISIC benchmarks and data augmentation techniques, while standing out for its strong emphasis on low computational cost alongside high predictive performance.

Targeting mpox detection, \cite{11129704}. present XceptMPX, a modified Xception model trained on MCVSLD, a newly compiled six-class dataset of 11,325 images expanded to 67,950 through augmentation, using an 80:20 split. Xception's classification head is replaced by Global Average Pooling followed by dense layers with Leaky ReLU, batch normalization, L2 regularization, and dropout, with the deepest 20 base layers fine-tuned using a cosine-annealing scheduler. Accuracy reached 96.36\% and 97.01\% on the original and augmented sets respectively, confirmed by 5-fold cross-validation at 97.06\%. A Streamlit-based web application supports real-time clinical deployment. Limitations include the absence of hospital-grade clinical validation, sensitivity to image quality and skin tone variation, and the preliminary scope of the usability evaluation.

\cite{11186509} propose XAI-SkinCADx, a six-stage deep ensemble framework applied to the ISIC dataset with 2,367 dermoscopic images across nine classes, expanded via augmentation to 5,023 balanced samples using an 80:20 split. Feature extraction combines LBP and GLCM descriptors with three DenseNet201-based CNN pairs followed by BiLSTM temporal modeling and Multiclass SVM classification. The best configuration, DM-25 with BiLSTM and SVM, achieved 95.63\% accuracy and 0.97 AUC, confirmed by 5-fold cross-validation. Explainability is addressed through Grad-CAM++ and LIME, with a sixth stage implementing a rule-based recommendation system stratifying predictions into low, moderate, and high clinical risk levels. Limitations include reliance on a single dataset without external validation and constraints in the representativeness of synthetic augmentation for real-world clinical scenarios.

Addressing binary skin cancer classification, \cite{11184754}, introduce EDA-ResNet50, trained on 3,297 dermoscopic ISIC images without augmentation using an 80:20 split. ResNet50 is enhanced with a Multi-Scale Feature Representation block and an Efficient Dual Attention module combining channel recalibration and spatial attention, initialized with ImageNet weights. The model reached 93.18\% accuracy and 94\% sensitivity, statistically validated through Cochran's Q and McNemar's tests, with generalizability demonstrated across HAM10000, a combined ISIC 2019/2020 dataset, and a blood cell benchmark. Limitations include the binary classification scope, absence of quantitative explainability evaluation against expert annotations, and dependence on dermoscopic equipment unavailable in low-resource settings.

\cite{11263785} propose CA Y-Net, a dual-branch CNN multitask learning architecture that integrates Convolutional Block Attention Modules (CBAMs) and a novel contrastive loss function (MuCo) to simultaneously perform skin lesion classification and segmentation. The model was evaluated on the International Skin Imaging Collaboration ISIC 2017 dataset, comprising 2,000 training images, 150 validation images, and 600 test images across three lesion classes, employing a hold-out strategy with an additional set of 1,036 images from the ISIC archive for extended training. Performance was assessed using mean AUC for classification and Jaccard Index for segmentation, achieving 92.8\% mean AUC and 79.6\% Jaccard Index, with additional validation on the Pedro Hispano Hospital PH2 dataset (98.9\% AUC and 91.3\% Jaccard), demonstrating good generalization capability. Although the results are competitive with the state of the art, the restriction of the study to a single challenge dataset and the requirement for segmentation annotations for all images limit conclusions regarding performance in broader and more heterogeneous clinical scenarios.

\cite{11284879} propose SynthraXCoreNet, an ensemble of six heterogeneous CNNs — ResNet50V2, ResNet101V2, ResNet152V2, DenseNet201, NASNetLarge, and Xception — combined through posterior probability soft voting, incorporating probability calibration via temperature scaling and an interpretability pipeline with Integrated Gradients, multiple Grad-CAM variants, and quantitative XAI metrics. The evaluation was conducted on four public datasets (HAM10000, ISIC2019, BCN20000, and DERM12345, totaling more than 50,000 images), using a stratified 70/20/10 split with intra-split class balancing. Performance was reported using accuracy, macro F1-score, and 95\% Wilson confidence intervals, with results of 98.85\% ± 0.68 on HAM10000 and 96.63\% ± 1.00 on DERM12345 (40 subclasses). Despite the strong performance and robust evaluation across multiple benchmarks, the absence of prospective multicenter validation with dermatologist approval and the inference latency of approximately 461 ms per image represent important limitations for real-world clinical deployment.

\cite{11354481} present GLR-Net, a hierarchical encoder-decoder architecture based on the Mix Vision Transformer (MiT) from SegFormer, integrating four synergistic components: Global-Local Refinement (GLoR), Boundary-Semantic Integration and Selection Filter (BISF), Attention-Gui-\\ded Context Refinement (AGCR) and multi-scale Reverse Attention (RA), to specifically enhance dermoscopic lesion boundary delineation. The model was evaluated on the ISIC 2016, 2017, 2018 and PH2 datasets, strictly following the official challenge splits (2.000/150/600 for ISIC 2017 and 2.594/100/1.000 for ISIC 2018), with images resized to 512×512 pixels and evaluated using Dice, IoU, accuracy, sensitivity, HD95 and ASSD. GLR-Net achieved 92.32\% Dice and 90.71\% IoU on ISIC 2017, as well as 95.99\% Dice on PH2. The cross-dataset generalization analysis, in which the model was trained on ISIC 2017 and tested on unseen datasets, including 500 samples from HAM10000 with 91.65\% Dice, strengthens the study’s conclusions. Nevertheless, the high computational demand of the transformer backbone (92.33M parameters and approximately 210 GFLOPs) and the lack of evaluation across skin tone subgroups limit the extension of the findings toward equitable clinical deployment.

\cite{11482135} conduct a controlled evaluation of six pre-trained CNN architectures: ResNet50, Xception, EfficientNetB3, MobileNetV2, Den-seNet201 and InceptionV3, along with their soft-voting ensemble for seven-class classification on the HAM10000 dataset, emphasizing architectural comparability and interpretability through Grad-CAM, LIME and Occlusion Sensitivity. The unified protocol employed a stratified train/test split with a fixed random seed (random seed = 42), inverse-frequency class weights to mitigate imbalance and evaluation using accuracy, F1-score, recall, precision, specificity and macro AUC-ROC. The ensemble achieved 89.37\% accuracy and a macro AUC of 0.985, outperforming the best individual model, ResNet50 (86.03\%). Although the controlled comparative approach represents a relevant methodological strength, the absence of cross-validation, the lack of guaranteed patient-wise separation, the absence of probability calibration and the strictly qualitative XAI analysis limit both the statistical robustness of the results and the formal assessment of the system’s clinical reliability.

The review of the analyzed articles revealed important insights into the use of deep learning for the classification and segmentation of dermatological lesions. Transfer learning with pre-trained architectures such as InceptionV3, Xception, and EfficientNet demonstrated high effectiveness, achieving accuracy values above 98\% on the HAM10000 dataset \citep{10608138,10416885}. Attention mechanisms and feature fusion strategies, as employed in models such as SPCB-Net and EFAM-Net, improved feature representation and contributed to enhanced classification performance \citep{10374026,10695064}. Data augmentation techniques, including conventional transformations and GAN-based image generation, were shown to mitigate class imbalance and improve model robustness \citep{10608138,10416885}. Overall, a predominance of transfer learning was observed, along with the recurrent use of the HAM10000 and ISIC datasets and the widespread adoption of data augmentation techniques, which contributed to the development of models with improved generalization capability.

Among the best practices identified, the use of pre-trained models and the combination of multiple evaluation metrics stood out, enabling a more comprehensive assessment of model performance. Ensemble strategies, such as those adopted in SynthraXCoreNet \citep{11284879} and the soft-voting ensemble evaluated by \citep{11482135}, consistently outperformed individual architectures, reinforcing their utility in imbalanced classification scenarios. Interpretability tools such as Grad-CAM and LIME, reported in several studies \citep{11482135,11284879}, also emerged as relevant complementary components for supporting clinical trust in model predictions.

The lack of standardization in evaluation metrics and data splitting strategies hindered direct comparison among the analyzed studies. Several works adopted hold-out splits with varying proportions, while others employed k-fold cross-validation, making performance comparisons across methods difficult to draw. Furthermore, reliance on publicly available imbalanced datasets, most notably HAM10000, compromises the generalization capability of networks for underrepresented classes, a limitation explicitly reflected in the modest macro recall values reported by studies such as \cite{11121145}. The absence of external or prospective validation was also a recurring limitation, noted in works such as \cite{10965692} and \cite{11284879}, restricting conclusions about real-world clinical applicability.


\cite{Alenezi2023} propose a multi-stage melanoma recognition framework that combines a hair-removal pre-processing step based on dilation and max-pooling with a pre-trained ResNet101 backbone for deep feature extraction, followed by Relief-based feature selection and a Bayesian-optimization-tuned SVM classifier. The model was evaluated on two balanced binary datasets built from ISIC-2019 and ISIC-2020, containing 1,168 and 9,044 dermoscopy images of melanoma and benign lesions, respectively, using an 80/20 hold-out split and accuracy, sensitivity, specificity, and precision as metrics. The full pipeline reached 99.14\% and 98.62\% accuracy on the two datasets, outperforming seven pre-trained backbones tested as baselines and previous state-of-the-art works such as Sayed et al. (2021, 98.37\%) and Pyingkodi et al. (2020, 98.32\%). Despite these strong results, the framework restricts the task to binary melanoma-versus-benign discrimination and adds non-negligible training overhead from the Bayesian hyperparameter search, as acknowledged by the authors themselves.

\cite{Hu2022} present AS-Net, an Attention Synergy Network for skin lesion segmentation that pairs a pre-trained VGG16 encoder with a decoder containing parallel spatial and channel attention paths, fused through a synergy module, and trained with a novel weighted binary cross-entropy loss to counter foreground-background size imbalance. Evaluated on ISIC2017 (2000/150/600 train/validation/test images), ISIC2018 (2594 images, 75/25 split), and PH2 (200 images, trained on ISIC2016), with online augmentation (random zoom, rotation, shifts) and results averaged over 15 runs, the model is assessed via pixel-wise Accuracy, Dice, Jaccard, Sensitivity, and Specificity. AS-Net attains 94.66\% ACC, 88.07\% Dice, and 80.51\% Jaccard on ISIC2017, and 95.68\% ACC/83.09\% Jaccard on ISIC2018, outperforming CDNN, FrCN, SLSDeep, DCL-PSI, and DA-Net, though the authors acknowledge the VGG16 backbone and dual attention paths inflate parameter count (24.9M) and computational cost relative to lighter architectures.

Complementing attention-based approaches to context modeling, \cite{Wang2022} introduces a cascaded context enhancement network for automatic skin lesion segmentation, combining a ResNet-ASPP encoder with a cascaded context aggregation (CCA) module that uses gate-based units to sequentially fuse original-image cues with multi-level encoder features, plus a context-guided local affinity (CGL) module that exploits this global context to refine local feature discrimination. Trained with a joint weighted binary cross-entropy and Dice loss and an auxiliary supervision branch, the model is evaluated on ISIC-2016, ISIC-2017, ISIC-2018 (five-fold cross-validation), and PH2, reaching Jaccard Index scores of 87.1\%, 80.3\%, 84.3\%, and 86.6\%, respectively, and Dice scores up to 92.6\% and 87.8\% on ISIC-2016 and ISIC-2017, outperforming CPFNet, Inf-Net, and PyDiNet by 1.1-1.5\% JA. However, the PH2 result stems from directly applying the ISIC-2017-trained model without fine-tuning on only 200 images, and the lack of an official ISIC-2018 test split limits direct comparability with challenge leaderboards.

\cite{Abdelhalim2021} propose SPGGAN-TTUR, a self-attention- augmented Progressive Growing GAN combined with the Two-Timescale Update Rule, designed to synthesize fine-grained 256×256 dermoscopic images for data augmentation rather than for direct classification. Evaluated on the HAM10000 dataset (10,015 images, 9514/501 train/held-out split across seven classes), the generated samples are used to retrain a ResNet-18 classifier under stratified 3-fold cross-validation. Using GAN-train/GAN-test protocols, SPGGAN-TTUR outperforms plain PGGAN and SPGGAN (68.1\% vs. 63.5\%/64.8\% GAN-train), and the resulting augmentation scheme raises overall sensitivity by 5.6\% over non-augmented training and 2.5\% over the best classical augmentation, with melanoma-specific recall improving by 13.8\% and 8.6\% respectively (p<0.05). However, absolute accuracy remains modest (66.1\%), and the authors themselves note that hardware constraints limited generation to 256×256 rather than the dataset's native 600×450 resolution, leaving open whether gains would persist at full resolution.

In a related effort targeting boundary precision and cross-dataset robustness, \citep{Zhu2025} propose EM-Net, a hybrid CNN-ViT encoder-decoder for skin lesion segmentation that couples a Morphology-aware Module—using non-convex optimization over a generalized M-junction field to extract fuzzy lesion boundaries—with Bridge Fusion Modules for multi-scale feature integration and a Few-shot Domain Generalization module that adapts a source-trained model to new target domains with minimal labeled samples. Evaluated on ISIC 2016/2017/2018, PH2, PAD-UFES-20, and the University of Waterloo dataset using Dice, IoU, Accuracy, Precision, and Recall, EM-Net surpasses U-Net, TransUNet, Swin-UNet, and ICL-Net across the board, reaching Dice/IoU of 91.89\%/85.57\% on ISIC 2016 and 94.03\%/88.92\% on PH2, and improving IoU over TransUNet by 5.62\% on Waterloo and 1.57\% on PAD-UFES-20, though the authors themselves note the model's high parameter count constrains deployment on resource-limited clinical hardware.

\cite{Li2024} propose DSEUNet, a lightweight U-shaped encoder-decoder network for skin lesion segmentation that combines four modules: Conditional Parameterized Convolution (CPC), Grouped Hadamard Product Attention (GHPA), a Dynamic Aggregation Bridge (DAB) replacing standard skip connections, and Spatial Group-Enhanced attention (SGE), trained with a multi-scale weighted BCE-Dice loss. Evaluated on ISIC2017 (2,150 images) and ISIC2018 (2,694 images) with a 7:3 train/test split, using mIoU and Dice Similarity Coefficient as metrics, the model attains 79.2\% mIoU / 88.4\% DSC on ISIC2017 and 80.5\% mIoU / 89.2\% DSC on ISIC2018, with only 23.306K parameters and 9.497M FLOPs, outperforming UNext-s by 1.6-1.8\% mIoU while using roughly half the parameters of EGEUNet. Despite this efficiency, DSEUNet still trails the much heavier TransUnet (85.9\%/85.1\% mIoU), suggesting an accuracy ceiling imposed by such aggressive parameter reduction.

In a related direction focused on segmentation rather than classification, \cite{Kaymak2020} conduct a comparative experimental study of four Fully Convolutional Network variants—FCN-AlexNet, FCN-8s, FCN-16s, and FCN-32s—for automatic skin lesion delineation on the ISIC 2017 dataset (2000 training, 150 validation, and 600 test dermoscopic images), leveraging transfer learning from ImageNet and PASCAL VOC pretraining. On the test set, FCN-8s achieved the best overall performance, reaching 93.9\% accuracy, 84.1\% Dice, and 72.5\% Jaccard, outperforming U-Net, LinkNet, SegNet, and II-FCN baselines and producing markedly sharper lesion contours than the coarser FCN-AlexNet and FCN-32s predictions. Despite these gains, the deeper FCN-8s/16s/32s models required roughly three times longer to train (over 500 minutes) than FCN-AlexNet (176 minutes), and the study neither applies data augmentation nor evaluates generalization beyond the single ISIC 2017 benchmark, limiting its practical scalability.

Extending the ensemble-learning paradigm to multi-scale feature fusion, \cite{Reis2026} propose PADSCNet, P-DEFF, and M-DEFF for skin cancer detection, where PADSCNet combines depthwise separable convolutions with multi-head attention across dual pathways (1175 layers but only 3.3 million parameters), while P-DEFF and M-DEFF fuse PADSCNet features—augmented by FCIE/CLAHE-enhanced images and, for M-DEFF, additional pre-trained backbones (DenseNet201, ResNet152, MobileNet)—via a hard-voting SVM-RF-KNN ensemble tuned with PSO/GSA/ GWO. On the Skin Cancer: Malignant vs. Benign dataset, PADSCNet, P-DEFF, and M-DEFF reach 88.12\%, 93.75\%, and 95.00\% accuracy respectively, with PADSCNet also generalizing to 93.12\% on the CNN for Melanoma Detection dataset and 95.00\% on PH2, outperforming heavier CNNs such as DenseNet201 (18.3M parameters) and ResNet152 (58.4M parameters). However, the PH2 result is inflated by severe class imbalance (only 8 melanoma test samples), where PADSCNet's melanoma sensitivity drops to 75\%, undermining the headline accuracy as a reliable clinical indicator.

Building on attention-based feature fusion strategies, \cite{Zhang2026} present GL-MSFN, a dermoscopic image classification framework built upon a customized EfficientNetV2-S backbone augmented with three complementary modules: an Adaptive Perception Aggregation Module (APAM) that reweights multi-scale features within MBConv blocks, a Global-Local Feature Fusion Module (GLFFM) with an asymmetric dual-branch attention design, and a Multi-Level Attention Module (MLAM) that replaces standard Squeeze-and-Excitation blocks with directional strip pooling, complemented by a combined focal and class-weighted loss to counter class imbalance. Trained on the eight-class ISIC 2019 dataset (25,331 images, 8:2 split), the model attains 93.85\% accuracy, 91.23\% precision, 92.70\% recall and 98.70\% specificity, surpassing DeepLabV3+, InSiNet and MetaFormers with focal self-attention, and generalizes to ISIC 2018 with 94.76\% accuracy, though cross-dataset evaluation remains confined to the closely related ISIC family rather than independent clinical cohorts.

\cite{Aruk2025} propose a hybrid CNN-ViT architecture for skin lesion classification that augments the MetaFormer-based CAFormer-s32 backbone with ConvNeXt blocks across a four-stage design (four ConvNeXt blocks in Stage 1, eight in Stage 2, sixteen transformer blocks in Stage 3, and four in Stage 4), using ConvNeXt's large-kernel depthwise convolutions and layer normalization for fine-grained local feature extraction while transformer blocks capture global context. Evaluated on HAM10000 and ISIC 2019 against ten CNN and ten ViT baselines under identical training conditions, the model, with only 38.01M parameters, achieves 94.30\% accuracy and 91.11\% F1-score on HAM10000, and 92.50\% accuracy and 90.38\% F1-score on ISIC 2019, with statistically significant gains confirmed via Wilcoxon signed-rank testing; however, performance still degrades noticeably on minority classes such as AK (F1 around 80\%), and both datasets predominantly represent light-skinned populations, limiting demonstrated generalizability across skin tones.

Departing from approaches that rely on transfer learning or heavy augmentation to compensate for scarce dermoscopic data, \cite{PezhmanPour2020} proposes a compact encoder-decoder CNN trained entirely from scratch for skin lesion and dermoscopic feature (globule/streak) segmentation, injecting multiscale, multidirectional contourlet-transform representations and CIELAB color channels directly into the pooling layers rather than deepening the network. On ISIC 2016 and 2017, the transform-domain variant lifted the Jaccard index by 12\% over the base 7-layer model, versus only 6\% from simply extending it to 15 layers, and the final model surpassed the ISIC 2017 challenge winner by 2.2\% Jaccard and 7\% sensitivity, and improved dermoscopic feature segmentation by 17\% Dice. However, the approach still depends on handcrafted transform features and manual post-processing heuristics, and validation is confined to the same two small ISIC challenge sets used by nearly every competing method.

\cite{Sethanan2023} propose a double artificial multiple intelligence system (AMIS)-ensemble deep learning framework for skin cancer classification, combining an ensemble image segmentation stage (thresholding, edge detection, region-growing, clustering, U-Net, and RP-Net fused via Jaccard similarity) with a heterogeneous ensemble of six CNN backbones (including DenseNet121, NASNetMobile, EfficientNetB7, EfficientNetV2L, and EfficientNetV2M) whose decisions are fused using AMIS-optimized weights, and the model is deployed through a Line chatbot for home use. Evaluated on HAM10000, a malignant-vs-benign dataset, and a merged HAM-MB-13312 dataset (80/20 split), the model reaches over 99.4\% accuracy on the aggregated data, 95.078\% accuracy through the chatbot, and outperforms prior CNN-based methods by 2.831-15.357\% on HAM10000 and 1.356-8.560\% on the malignant-vs-benign dataset, with a reported System Usability Scale score of 96.85 from 31 clinical volunteers. The heavy reliance on a computationally expensive six-CNN ensemble with metaheuristic weight optimization raises concerns about training cost and real-time deployability despite the strong reported accuracy gains.

In a related direction, \cite{Sulthana2024} propose an end-to-end deep convolutional neural network framework, S-MobileNet, for multiclass skin lesion classification on the HAM10000 dataset (10,000 dermoscopic images, 7 classes, 80:20 train-test split), combining a custom Gaussian-based segmentation algorithm with a modified Segmentation-based Fractal Texture Analysis (SFTA) feature extractor before classification. The architecture is evaluated along four axes: raw versus preprocessed input, ReLU versus Mish activation, and pruned versus unpruned layers, across Adam, RMSProp, and SGD optimizers. Their best configuration (preprocessed data, SGD, Mish, with L1-norm pruning of 156 filters) reached 98.35\% training and 98.15\% test accuracy, 96.23\% precision, and 94.59\% F1-score, reportedly surpassing MobileNet, ShuffleNet, DenseNet, and other HAM10000 benchmarks from 2018-2023. However, the comparison relies on figures copied from prior papers rather than re-implementation under identical hardware/latency conditions, weakening the validity of the superiority claim.

\cite{Gao2026} propose AMST-Net, an adaptive multi-scale transformer dual encoder network for skin lesion segmentation that pairs a light-\\weight, SegFormer-style hierarchical Transformer branch producing four coar-\\se-to-fine resolution features with a parallel CNN encoder, hierarchically fusing them through a cascaded multi-scale pooling efficient channel attention module (CMSP-ECA) and an efficient contextual information fusion module (CIFM) inside a U-shaped architecture trained with a combined Dice and cross-entropy loss. Evaluated on ISIC2017, ISIC2018, and PH2 with simple augmentation (Gaussian noise, flips, translation), the model reaches \\Dice/IoU/ACC of 88.53\%/81.48\%/92.07\% on ISIC2018, 84.32\%/75.79\%/\\82.66\% on ISIC2017, and 92.89\%/87.42\%/96.34\% on PH2, outperforming U-Net, CE-Net, CA-Net, FAT-Net, H-Net, TransUNet, TransFuse, and CMM-Net across Dice, IoU, ASSD, and HD95. Despite the consistent gains, the 32M-parameter model remains heavier than several competing baselines such as CA-Net (2.8M) and SegFormer (7M), and the authors themselves note persistent failures on extremely low-contrast or spatially discontinuous lesions.

Moving outside the dermoscopic domain entirely, \cite{Xu2024} propose DKNet, a domain knowledge-driven encoder-decoder for nasopharyngeal carcinoma (NPC) segmentation that encodes radiologists' expert-prior knowledge of NPC predilection sites as a Gaussian mixture distribution via optimal-transport regularization, combined with dynamic top-k pooling for slice-level prediction and a cross-scale feature refinement module for coarse-to-fine segmentation. Built on a modified ResNet-14 backbone (2.8M parameters), the model is evaluated on a private multi-hospital MRI dataset (SegNPC, 1,181 patients) and the public SegRap2023 CT dataset (120 patients) using AUC and Dice as metrics. DKNet reaches 98.2\% and 94.3\% AUC for primary-tumor and lymph-node prediction, respectively, and 85.5\% Dice for segmentation, outperforming AlexNet, ResNet, ViT, Swin, U-Net, and NPCNet while using only 10\% of the pixel-level annotations required by competing methods. However, the work targets an entirely different cancer type and imaging modality than dermoscopic skin lesion analysis, limiting its relevance to the domain-knowledge strategy itself.

In a further departure from dermoscopic classification, \cite{Huang2026} present M2CR, a multimodal diagnostic system for primary liver cancer subtypes (HCC, ICC, cHCC-CCA) that couples a segmentation subnetwork, ResUNetX, with a multimodal classification network, MCE-CCLNet, fusing CECT image features and laboratory indicators through cross-modal attention, a dynamic contrastive loss, and Intra-/Inter-phase BiLSTM modules for variable slice counts and phase combinations. Evaluated on a private multi-hospital 3D CECT dataset (361 subjects) and externally validated on the public LiTS dataset, ResUNetX outperforms U-Net, U-Net++, and SAM in IoU/Dice, and MCE-CCLNet reaches a macro-average accuracy of 97\% and AUROC of 96\%, outperforming MulT, MedFuse, and MMAformer, though performance drops to 86\% accuracy under leave-one-hospital-out cross-validation. The system was further piloted prospectively in a clinical setting integrated with hospital LIS/PACS infrastructure. As with \cite{Xu2024}, however, the work addresses a different cancer type and imaging modality entirely outside the dermoscopic domain, limiting its relevance to the multimodal fusion strategy it demonstrates.

As a guideline for this research, rigorously standardized evaluation protocols were adopted, including multiple performance metrics such as accuracy, precision, recall, F1-score, and AUC, to ensure a comprehensive assessment of model performance. Standardized preprocessing, dataset splitting, and class balancing strategies were also implemented to enable fair and reproducible comparisons.
The main differential of this work in relation to the state of the art lies in the proposal of a unified and reproducible experimental framework for skin cancer classification. Unlike previous studies that typically evaluate a single architecture or employ non-standardized validation protocols, this work performs a systematic and controlled comparison across multiple convolutional neural network architectures under identical experimental conditions. These contributions help address key limitations identified in prior studies, supporting the development of more robust, transparent, and clinically applicable artificial intelligence models for skin cancer diagnosis.

\section{Materials and Methods}
\label{sec:Materials-and-Methods}

\subsection{Datasets}
\label{sec:Datasets}

In this study, three distinct publicly available datasets were used, comprising both dermoscopic and histopathological images. Access information for all datasets is provided in the Data Availability Section~\ref{sec:dataavailability}. The HAM10000 and ISIC 2018 datasets consist of dermoscopic images, whereas the CR-AI4SkIN dataset corresponds to histopathological images derived from microscopic sections of skin tissue. In order to maintain a consistent experimental setting across datasets, the classification task was formulated as a binary problem, distinguishing melanoma (malignant) from benign (non-melanoma) lesions. The selection of these datasets enabled the evaluation of model robustness across different imaging modalities while maintaining the focus on melanoma detection.\subsubsection{HAM10000}

The HAM10000 dataset is one of the most widely used public datasets in skin lesion detection studies. It contains 10,015 high-resolution dermoscopic images distributed across seven original diagnostic classes \citep{DVN/DBW86T_2018}: \textit{akiec} (actinic keratoses and intraepithelial carcinoma), \textit{bcc} (basal cell carcinoma), \textit{bkl} (benign keratosis-like lesions), \textit{df} (dermatofibroma), \textit{nv} (melanocytic nevi), \textit{vasc} (vascular lesions), and \textit{mel} (melanoma). The dataset presents a significant class imbalance, with melanocytic nevi representing the majority of samples, while classes such as dermatofibroma and vascular lesions contain fewer instances.

For the purposes of this study, these categories were reorganized into two classes in order to standardize the binary classification task. The melanoma class includes only the \textit{mel} category, while the non-melanoma class includes all remaining categories (\textit{akiec}, \textit{bcc}, \textit{bkl}, \textit{df}, \textit{nv}, and \textit{vasc}). This grouping allows the models to focus specifically on the distinction between melanoma and other skin lesion types.

The images were acquired under varying lighting conditions and using different dermoscopic devices, resulting in variations in color, contrast, and resolution. These characteristics make the HAM10000 dataset particularly suitable for evaluating the robustness and generalization capability of deep learning models in dermoscopic image classification.

\subsubsection{ISIC 2018}

The ISIC 2018 dataset (Task 3: Lesion Diagnosis), released as part of the International Skin Imaging Collaboration (ISIC) Challenge \citep{codella2019skin}, contains dermoscopic images annotated for lesion diagnosis. In this study, the Training Ground Truth subset was used, comprising 10,015 images distributed across seven diagnostic categories: \textit{mel} (melanoma), \textit{nv} (melanocytic nevus), \textit{bcc} (basal cell carcinoma), \textit{akiec} (actinic keratosis and intraepithelial carcinoma), \textit{bkl} (benign keratosis-like lesions), \textit{df} (dermatofibroma), and \textit{vasc} (vascular lesions). Similar to other dermoscopic datasets, the class distribution is imbalanced, with melanocytic nevi representing the majority of samples.

The images were acquired in different clinical settings using multiple dermoscopic devices, resulting in variations in illumination, color distribution, resolution, and lesion appearance. These characteristics make the dataset particularly suitable for evaluating the robustness and generalization capability of deep learning models.

For the purposes of this study, the original diagnostic categories were reorganized into two classes. The melanoma class includes only the \textit{mel} category, while the non-melanoma class includes all remaining categories (\textit{nv}, \textit{bcc}, \textit{akiec}, \textit{bkl}, \textit{df}, and \textit{vasc}). This grouping standardizes the binary classification process and allows the models to focus on the distinction between melanoma and other skin lesion types.

The ISIC 2018 dataset is widely used as an international benchmark for automated skin lesion classification, reinforcing its relevance for validating and comparing deep learning approaches.

\subsubsection{CR-AI4SkIN}

The CR-AI4SkIN dataset consists of histopathological image samples collected from 215 patients, with annotations provided by expert dermatopathologists and multiple non-expert annotators \citep{del2023annotation}. Each case is associated with several image patches extracted from whole-slide images, representing different tissue regions and morphological structures. This patch-based organization increases the variability within each case and reflects realistic histopathological analysis conditions.

The dataset includes binary diagnostic labels corresponding to melanoma and non-melanoma. In addition to the expert ground truth (GT), the dataset also provides majority vote (MV) labels and individual annotations from multiple non-expert annotators. However, for the purposes of this study, only the expert annotations provided by dermatopathologists (ground truth – GT) were used for training and evaluation, ensuring that the models were supervised exclusively using clinically validated diagnoses.

Unlike dermoscopic datasets such as HAM10000 and ISIC 2018, CR-AI4SkIN contains histopathological images, which present substantially different visual characteristics, including cellular-level structures, staining patterns, and complex tissue morphology. These differences make the classification task more challenging and require the models to learn more discriminative and robust feature representations.

This dataset was incorporated to evaluate the generalization capability of the models across different imaging modalities and to assess their performance in a more complex and clinically detailed diagnostic scenario.

\subsection{Data Preprocessing}
\label{sec:Data-Preprocessing}

Images from all datasets were resized to a fixed spatial resolution of 224×224 pixels, converted to RGB format, and normalized to ensure consistency in the input dimensions of the convolutional neural networks. The 224×224 resolution was adopted because it corresponds to the standard input size required by most pretrained CNN architectures evaluated in this study, allowing a consistent comparison while preserving sufficient image detail for lesion characterization. For the InceptionV3 architecture, images were resized to 299×299 pixels, following the original input specification of the model. This standardization step guarantees a uniform spatial representation across datasets acquired under different imaging conditions and devices.

To mitigate class imbalance and increase intra-class variability, data augmentation was performed using the Albumentations library \citep{albumentations2025}. The augmentation pipeline included horizontal flipping (p=0.5), random brightness and contrast adjustment (p=0.2), random rotation (limit $\pm 30^\circ$, p=0.5), affine transformations with scaling between 0.9 and 1.1, translation of up to 5\%, additional rotation (p=0.5), Gaussian blur (p=0.1), and random resized cropping with scaling between 80\% and 100\% of the original image size (p=0.5). These transformations were applied to generate synthetic samples for underrepresented classes until a predefined target number of images per class was reached, resulting in balanced class distributions. This strategy aimed to increase variability, reduce overfitting, and strengthen generalization capability.

For the HAM10000 dataset, the balancing procedure resulted in 3000 samples per class (melanoma and non-melanoma), totaling 6000 images. For the ISIC 2018 dataset, 1800 samples per class were used, totaling 3600 images. After balancing, the original multiclass labels were converted into a binary classification problem, in which melanoma samples were grouped into the malignant class and all remaining lesion types were grouped into the benign (non-melanoma) class.

The datasets were split into 70\% for training, 15\% for validation, and 15\% for testing in the case of HAM10000 and ISIC 2018, ensuring a balanced distribution for model development and evaluation. For the CR-AI4SkIN dataset, the split followed the original recommendation provided by the dataset authors, with 150 patients allocated for training, 23 for validation, and 43 for testing. This patient-level separation prevented data leakage and ensured a consistent and reliable performance evaluation protocol. Notably, the number of images per patient is not standardized, varying according to the specific area analyzed; consequently, the volume of data per patch ranges significantly, from approximately 60 images to over 1,000 crops.

\subsection{Pre-trained Models}
\label{sec:Pre-trained-Models}

The architectures evaluated in this study include ResNet50, MobileNet, VGG16, VGG19, and InceptionV3, each selected based on specific architectural strengths demonstrated in recent skin lesion classification research. ResNet50 employs residual connections that enable the training of deeper networks without degradation, improving feature extraction and classification performance in dermoscopic datasets \citep{Akter_2022}. MobileNet employs depthwise separable convolutions that significantly reduce computational complexity and the number of parameters while preserving effective feature extraction capability, making it suitable for efficient skin lesion image classification tasks \citep{Howard2017}. VGG16 employs a deep architecture composed of sequential convolutional layers with small receptive fields, enabling progressive hierarchical feature learning and effective extraction of visual patterns relevant for accurate skin lesion classification \citep{Simonyan2015}. VGG19 extends this capability with increased depth, enabling improved abstraction of high-level features and more detailed feature representations, which enhances classification robustness in complex dermatological image analysis tasks \citep{Al_Shafi_2025}. InceptionV3 utilizes multiple convolutional filter sizes within its Inception modules, enabling the model to capture both local and global features, which improves its ability to differentiate between various types of skin lesions and enhances classification accuracy in dermatological image analysis \citep{Jayanti2024}. These architectures are widely used in medical image classification tasks and are recognized for their effectiveness in extracting complex visual features such as lesion shape, border irregularities, texture patterns, color variations, and structural asymmetries, which are critical for accurate skin lesion analysis and discrimination between benign and malignant cases. Unlike traditional approaches where CNNs act as end-to-end classifiers, in this work these architectures were used as feature embedding extractors, generating feature vectors that were integrated into a supervised metric learning framework to improve class separability. Each model was initialized with weights pre-trained on the ImageNet dataset \citep{Deng2009}, which contains tens of millions annotated images across thousands of object categories. This strategy enables knowledge transfer, accelerates convergence, and reduces the risk of overfitting when dealing with limited medical image datasets.

The feature embeddings generated by each network were organized within an optimized metric space. Activations from intermediate layers were converted into feature vectors, which were then used as input to a distance-based classifier, namely k-NN (k-Nearest Neighbors). This supervised method assigns a label to a new sample by identifying, in the embedding space, the k-nearest examples according to a similarity measure, such as Euclidean distance \citep{hu2016distance}. This approach enables the evaluation of the discriminative quality of the learned feature representations while separating the feature extraction process from the final classification stage. In addition, k-NN was selected because it operates directly in the embedding space without introducing additional trainable parameters, allowing a more reliable assessment of the intrinsic quality and separability of the learned feature representations.

To ensure a fair and controlled comparison among the evaluated architectures, a standardized training protocol was established and applied consistently to all models. This protocol included fixed hyperparameters such as learning rate, batch size, optimizer configuration, loss function, and embedding dimensionality, as well as the same projection head structure and training strategy. By maintaining these parameters constant across all networks, the experimental design ensures that performance differences can be attributed primarily to the backbone architecture rather than variations in training configuration.

\subsubsection{ResNet50 Architecture}
\label{sec:resnet50-architecture}
\begin{figure}[t]
\centering
\includegraphics[width=1.0\textwidth]{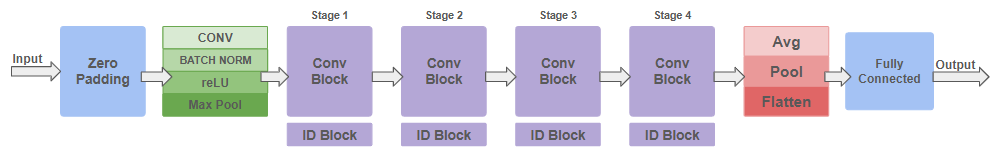}
\caption{ResNet50 architecture overview comprising zero padding, four residual stages, average pooling, and fully connected output layer. Source: Author, based on \citep{mukherjee2022resnet50}.}
\label{fig:resnet50_architecture}
\end{figure}

ResNet50 enables the training of deep neural networks with 50 convolutional layers through the use of skip connections, which preserve information flow across layers and mitigate the vanishing gradient problem, thereby facilitating the learning of complex and hierarchical feature representations \citep{pytorchresnet50}as illustrated in \autoref{fig:resnet50_architecture}. By incorporating residual learning, the architecture improves gradient propagation during backpropagation, allowing deeper structures to be optimized more effectively and consistently. This design enhances convergence stability and supports the extraction of high-level semantic features from input images. Due to its depth, robustness, and strong representational capacity, ResNet50 has become a widely adopted baseline model for feature representation learning and for comparative evaluation with other convolutional neural network architectures in image classification tasks.

\subsubsection{MobileNet Architecture}
\label{sec:mobilenet-architecture}
\begin{figure}[t]
\centering
\includegraphics[width=1.0\textwidth]{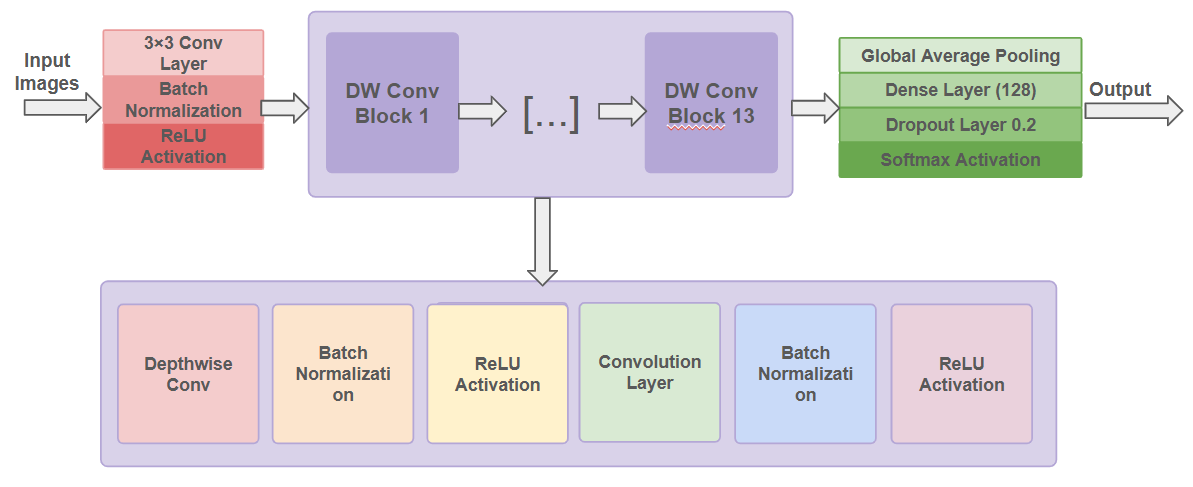}
\caption{MobileNet architecture overview comprising an initial convolution block,
13 depthwise separable convolution blocks, global average pooling, and softmax output.Source: elaborated by the author based on \citep{asif2024lwse}.}\label{fig:mobilenet_architecture}
\end{figure}

MobileNet was designed with a focus on efficiency and portability, ensuring a balance between performance and computational cost. Its architecture is primarily based on depthwise separable convolutions, as illustrated in \autoref{fig:mobilenet_architecture}, which decompose the standard convolution into a depthwise convolution (applied independently to each input channel) followed by a pointwise 1×1 convolution responsible for channel combination. This factorization significantly reduces the number of parameters and computational operations compared to conventional convolutions, while preserving discriminative capacity \citep{mobilenets}. Additionally, MobileNet introduces width and resolution multipliers that allow control over the trade-off between accuracy and efficiency, making it particularly suitable for deployment on resource-constrained devices while maintaining competitive performance in image classification tasks.

\subsubsection{VGG16 and VGG19 Architectures}
\label{sec:vgg-architecture}
\begin{figure}[t]
\centering
\includegraphics[width=0.6\textwidth]{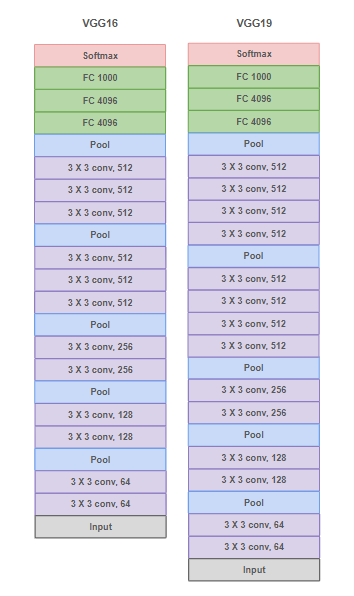}
\caption{VGG16 and VGG19 architecture overview comprising stacked $3\times3$
convolutional layers, max-pooling stages, and fully connected layers with softmax output.
Source: elaborated by the author based on \citep{datahacker2018vgg}.}
\label{fig:vgg_architecture}
\end{figure}

VGG16, developed by the Visual Geometry Group at the University of Oxford \citep{app112311185}, consists of 16 weight layers organized in a sequential architecture that stacks small $3 \times 3$ convolutional filters with stride 1, each followed by ReLU activation functions, and periodically applies $2 \times 2$ max-pooling layers to progressively reduce spatial dimensions, as illustrated in \autoref{fig:vgg_architecture}. This uniform design increases network depth while keeping the convolutional operations simple and consistent across layers. As the network deepens, it captures increasingly abstract and hierarchical visual patterns, transitioning from low-level features such as edges and textures to high-level semantic representations. Despite its relatively high number of parameters, VGG16’s straightforward and homogeneous architecture has made it a widely adopted baseline and reference model in image classification tasks.

VGG19, a deeper variant of VGG16, incorporates additional convolutional layers, increasing its representational and learning capacity \citep{manataki2023}. Maintaining the same design principle of stacked $3 \times 3$ convolutional filters followed by ReLU activations and periodic $2 \times 2$ max-pooling operations, VGG19 extends the network depth to 19 weight layers. This increased depth enables the extraction of more detailed and fine-grained hierarchical features, enhancing the model’s ability to discriminate subtle visual patterns. Such capability is particularly valuable in medical image classification tasks, including the differentiation between benign and malignant dermatological lesions.

\subsubsection{InceptionV3 Architecture}
\label{sec:inceptionv3-architecture}
\begin{figure}[t]
\centering
\includegraphics[width=1.0\textwidth]{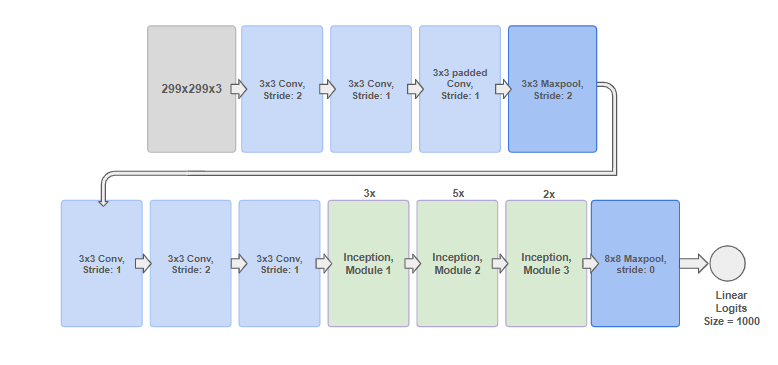}
\caption{InceptionV3 architecture overview comprising initial convolution stages,
three inception module groups, and a linear output layer.
Source: elaborated by the author based on \citep{asif2024lwse}.}
\label{fig:inceptionv3_architecture}
\end{figure}

InceptionV3 was selected due to its efficient and modular architecture, which combines convolutions of different kernel sizes within parallel branches in the same block, as illustrated in \autoref{fig:inceptionv3_architecture}, allowing the network to capture multi-scale spatial features simultaneously. By factorizing larger convolutions into smaller operations and incorporating dimensionality reduction strategies, the model improves computational efficiency while maintaining strong representational capacity. This architectural design supports robust feature extraction and high classification performance, making InceptionV3 well suited for applications that require both efficiency and reliability, such as automated diagnostic systems \citep{Admass2024}.

\subsection{Training and Hyperparameter Settings}

The backbone architectures ResNet50, MobileNet, VGG16, VGG19, and InceptionV3 were initialized with ImageNet pre-trained weights. The convolutional base of each network was configured with \texttt{include\_top=False} and Global Average Pooling. All backbone layers were frozen during training, allowing only the embedding head to be optimized.

The embedding head consisted of a Dropout layer (rate = 0.5), followed by a fully connected layer with 256 neurons and ReLU activation, Batch Normalization, a second Dropout layer (rate = 0.3), and a Dense layer with 128 units. The final embeddings were L2-normalized to constrain the feature space.

Metric learning was performed using Triplet Loss with a margin of 0.5. The loss function computes the squared Euclidean distance between anchor-positive and anchor-negative pairs, enforcing a minimum margin between them. The Adam optimizer was used with a learning rate of $1 \times 10^{-4}$.

Triplets were dynamically generated during training. For each batch, anchor and positive samples were randomly selected from the same class, while negatives were sampled from different classes. When enabled, hard negative mining was applied by selecting the negative sample with the smallest embedding distance among a subset of candidates.

Data augmentation was applied using ImageDataGenerator with random rotations (up to 30 degrees), width and height shifts (0.1), shear transformations (0.1), zoom (0.1), horizontal flipping, and reflective padding.

Training was performed for up to 30 epochs using EarlyStopping with a patience of 15 epochs, restoring the best model weights based on validation loss.

Table~\ref{tab:hyperparameters_triplet} summarizes the hyperparameters used in the experiments.

\begin{table}[ht]
\centering
\caption{Hyperparameters used for training the evaluated architectures with Triplet Loss.}
\label{tab:hyperparameters_triplet}
\begin{tabular}{l c}
\hline
\textbf{Component} & \textbf{Configuration} \\
\hline
Pre-trained weights & ImageNet \\
Backbone trainable & No (Frozen) \\
Pooling strategy & Global Average Pooling \\
Embedding dimension & 128 \\
Dense layer & 256 (ReLU) \\
Batch Normalization & Yes \\
Dropout (1st layer) & 0.5 \\
Dropout (2nd layer) & 0.3 \\
Loss function & Triplet Loss \\
Triplet margin & 0.5 \\
Optimizer & Adam \\
Learning rate & $1 \times 10^{-4}$ \\
Batch size & 8 \\
Epochs & 30 \\
EarlyStopping patience & 15 \\
Data augmentation & Rotation, shift, shear, zoom, flip \\
Hard negative mining & Optional \\
\hline
\end{tabular}
\end{table}

In total, 15 models were constructed and evaluated in this study, corresponding to five backbone architectures (ResNet50, MobileNet, VGG16, VGG19, and InceptionV3) trained independently on three datasets: HAM10000, ISIC 2018, and CR-AI4SkIN. For each dataset, five distinct models were implemented, ensuring a consistent and comparable experimental protocol across all data sources.

\subsection{Model Evaluation}
\label{sec:Evaluation-Metrics}

The evaluation relied on traditional quantitative classification metrics combined with complementary analyses focused on embedding interpretation, training stability, and statistical comparison between models, enabling a comprehensive assessment of both discriminative capacity and result consistency.

These measures were computed based on the confusion matrix, which organizes model predictions by comparing predicted against true labels into four entries: true positives (TP) and true negatives (TN), representing correctly classified samples, and false positives (FP) and false negatives (FN), representing misclassifications from which all subsequent metrics are derived \citep{hicks2022evaluation}.

For model evaluation, the following measures were considered:

\begin{itemize}

\item \textbf{Accuracy}: measures the proportion of correct predictions relative to the total number of evaluated samples, offering an overall view of model performance \citep{Ghanem2023}. However, as medical datasets often present class imbalance, accuracy alone may not be sufficient to fully characterize model behavior.
\begin{equation}
\text{Accuracy} = \frac{TP + TN}{TP + TN + FP + FN}
\end{equation}

\item \textbf{Precision (per class)}: evaluates the proportion of true positives among all samples predicted as belonging to a given class \citep{Alnaggar2024}. Precision was computed separately for each class. In the binary analysis, the positive class corresponds to malignant lesions (\textit{e.g.}, melanoma and its variations), while the negative class corresponds to other lesion types.
\begin{equation}
\text{Precision} = \frac{TP}{TP + FP}
\end{equation}

\item \textbf{Recall (per class)}: also referred to as sensitivity, quantifies the proportion of actual samples of a given class that were correctly identified \citep{hicks2022evaluation}. Recall was calculated individually for each class, with particular emphasis on the malignant class due to its clinical relevance.
\begin{equation}
\text{Recall} = \frac{TP}{TP + FN}
\end{equation}

\item \textbf{F1-score (per class)}: calculated as the harmonic mean between precision and recall \citep{Ahmad2025}, and computed separately for each class. This metric provides a balanced assessment of classification performance, especially in the presence of class imbalance.
\begin{equation}
\text{F1-Score} = \frac{2 \cdot \text{Precision} \cdot \text{Recall}}{\text{Precision} + \text{Recall}}
\end{equation}

\item \textbf{ROC curve and Area Under the Curve (AUC)}: used to analyze the discriminative capability of the models across different decision thresholds \citep{li2024auc}. The AUC value quantifies the overall ability of the model to distinguish between malignant and non-malignant lesions, with values closer to 1 indicating better discrimination.
\begin{equation}
\text{AUC} = \int_{0}^{1} \text{TPR}(t)\, d\text{FPR}(t)
\end{equation}
where $\text{TPR}$ denotes the true positive rate (Recall), $\text{FPR}$ denotes the false positive rate, and $t$ represents the decision threshold.

\end{itemize}

\subsection{Experimental Settings}
\label{sec:Overall-Method-Workflow}

\begin{figure}[t]
\centering
\includegraphics[width=1.0\textwidth]{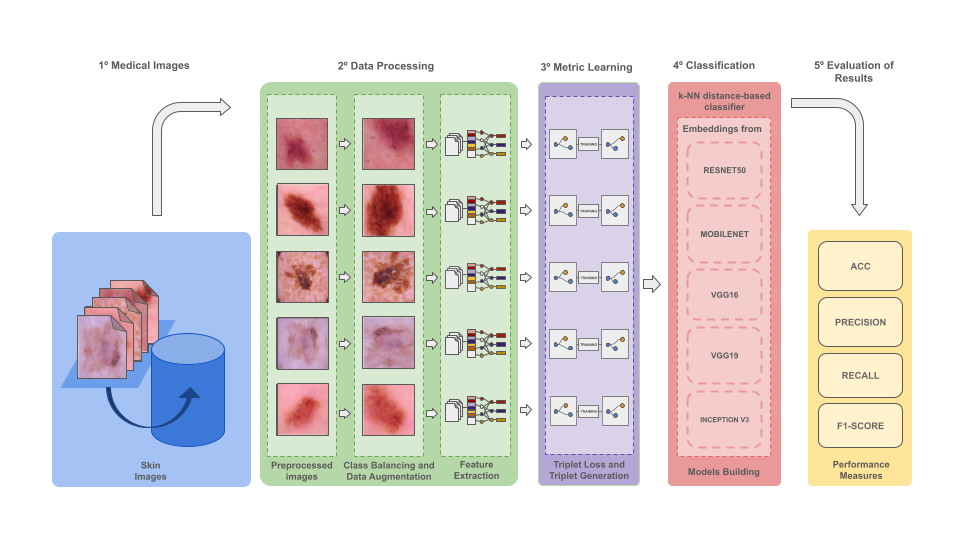}
\caption{Overall workflow of the proposed method. In the case of the CR-AI4SkIN dataset, pre-computed embeddings were directly used in the classification stage, and the image processing and metric learning steps were not applied.}
\label{fig:experimental_setup}
\end{figure}

The developed setup was structured into four stages, as illustrated in \autoref{fig:experimental_setup}. 

After image acquisition, resizing was performed according to the input size required by each CNN architecture, ensuring visual uniformity for processing and avoiding inconsistencies arising from the use of original image resolutions. For the HAM10000 and ISIC 2018 datasets, preprocessing was applied directly to each image file, while the corresponding CSV files were used exclusively to associate each image with its respective class label.

In contrast, for the CR-AI4SkIN dataset, this step was not performed, as the dataset was integrated into the pipeline through a previously prepared PKL file (a file format native to Python used to save and load precomputed data objects) containing embeddings organized by clinical case.

To mitigate issues related to class imbalance—which can compromise the discriminative capability of the models and introduce bias during training—a strategy combining sampling and data augmentation was adopted. The Albumentations framework\footnote{\url{https://albumentations.ai/}} was employed to expand minority classes until predefined target quantities were reached for each dataset. The applied transformations included horizontal flipping, brightness and contrast adjustments, controlled rotations, affine transformations with slight geometric variations, Gaussian blur, and random resized cropping (RandomResizedCrop), producing sufficient intra- and inter-class variability to enrich the sample distribution.

For the HAM10000 dataset, each minority class was expanded to 500 images, while the melanoma class was increased to 3000 samples. The strategy consisted of retaining the original images and generating augmented samples when necessary until the target size was reached. This configuration enabled the construction of a balanced binary dataset (Benign/Malignant), resulting in 3000 instances per class.

In the ISIC 2018 dataset, where the original distribution is even more asymmetric, the same augmentation pipeline was applied. All classes were balanced to 300 samples, except for the melanoma-related class, which was expanded to 1800 images. As with HAM10000, a binary version of the dataset was created, equally balanced with 1800 malignant and 1800 benign samples.

For the CR-AI4SkIN dataset, class balancing was not performed through data augmentation, since the dataset contains multiple image patches associated with each patient and provides preprocessed embeddings supplied by the repository. Therefore, the dataset expansion was performed only at a structural level by mapping all patches corresponding to each clinical case.

After the balancing step, each dataset was processed individually, maintaining separate workflows according to their specific characteristics. For the HAM10000 and ISIC 2018 datasets, all balanced images were processed by ImageNet-pretrained architectures, whose embeddings were refined through supervised metric learning using Triplet Loss. In this stage, the weights of the convolutional backbone were kept frozen, and only the layers responsible for embedding projection were trained, with the objective of organizing the vector space according to class similarity, maximizing inter-class separation and reducing intra-class variability. 

Metric learning is a representation learning paradigm that focuses on learning an embedding space in which the distance between samples reflects their semantic similarity \citep{musgrave2020metric}. In this paradigm, the model is trained to learn feature representations that organize the embedding space according to similarity relationships, encouraging samples that share semantic characteristics to be positioned closer together while separating dissimilar samples. By learning representations based on relative distances rather than explicit class decision boundaries, metric learning enables models to capture structured relationships among data samples and produce embeddings that are more suitable for similarity-based analysis and downstream classification tasks.

This approach is particularly advantageous in medical imaging tasks, where high intra-class variability and visual similarity between pathological and non-pathological patterns can make traditional classification boundaries difficult to learn \citep{Zeng2022}. By enforcing intra-class compactness and inter-class separation in the feature space, metric learning encourages the model to extract representations that are shared within each class but discriminative between classes, producing embeddings more suitable for downstream classification tasks. 

For the CR-AI4SkIN dataset, a different procedure was adopted. No new embedding extraction or Triplet Loss optimization was performed because the dataset provides precomputed feature vectors in a PKL file rather than the original image set. Therefore, these feature vectors were used directly as input to the classifier. This approach follows the format in which the dataset is publicly distributed and should be considered when interpreting the results obtained for this dataset.

The embeddings obtained after processing were used as input to a traditional classifier, specifically the K-Nearest Neighbors (KNN) algorithm. The classifier was configured with k=5 neighbors and Euclidean distance as the similarity metric, where k=5 was adopted as a widely used default value that balances sensitivity to local noise and preservation of local structure without requiring task-specific tuning. KNN was selected because it operates directly on feature similarity and does not require additional parameter learning, allowing the evaluation to focus on the discriminative quality of the learned embeddings. Since metric learning explicitly structures the feature space according to distance relationships, similarity-based classifiers such as KNN are particularly suitable for evaluating how well samples from the same class cluster together in the embedding space.

The classifier was trained directly on the embeddings generated by the evaluated architectures, including ResNet50, MobileNet, VGG16, VGG19, and InceptionV3, whose feature representations were optimized through Triplet Loss–based metric learning. In this methodology, the network learns to structure the embedding space using triplets composed of an anchor sample, a positive sample from the same class, and a negative sample from a different class. The optimization objective encourages the anchor representation to be closer to the positive sample than to the negative one by at least a predefined margin, promoting greater inter-class separation and reduced intra-class variability in the learned feature space.

For the CR-AI4SkIN dataset, the classifier was trained directly on the feature embeddings provided in the dataset repository through a PKL file. These embeddings were generated by the dataset authors using their own representation learning pipeline; therefore, no additional feature extraction or metric learning stage was applied in this work. Only the expert-defined ground-truth labels were used for training and evaluation, while the majority vote and individual annotator labels from the crowdsourcing process were not considered in the classification stage.

Performance evaluation was conducted using metrics derived from the confusion matrices of the predicted labels. After classification, accuracy and macro F1-score were computed to assess model performance. The classification task was formulated as a binary problem (benign vs. malignant) in order to emphasize the clinical relevance of early detection of malignant skin lesions, which is a critical objective in dermatological diagnosis.

\section{Results and Discussion}
\label{sec:Results-and-Discussion}

The experiments were conducted using five convolutional architectures pre-trained on ImageNet. The networks employed are ResNet50, VGG16, VGG19, MobileNet, and InceptionV3, which were used as embedding extractors within a supervised metric learning pipeline. The purpose of this stage was to transform each image into a high-dimensional vector representation capable of capturing discriminative properties relevant to the classification task. These representations served as input to the classifier used in the final implementation, ensuring consistency across all evaluated scenarios.

Subsequently, the embedding space was optimized through supervised metric learning using Triplet Loss, promoting greater inter-class separation and reduced intra-class variability, thus favoring a more separable structure suitable for the subsequent classifier. This process was applied only to the HAM10000 and ISIC 2018 datasets, which underwent the complete visual processing pipeline. In the case of CR-AI4SkIN, Triplet Loss was not applied, since the dataset already provides precomputed feature embeddings.

The models were then evaluated across the three datasets, enabling the analysis of the behavior of the evaluated architectures across different data scenarios. The results revealed variations among the architectures in terms of accuracy and macro F1-score (Table \ref{tab:acc-f1score}). In addition to these metrics, precision and recall values were also analyzed to provide a more detailed view of the classification behavior, particularly regarding the balance between correctly predicted cases and the detection of relevant lesions (Table \ref{tab:precision-recall}).

\begin{table}[h!]
\centering
\begin{tabular}{lccc}
\hline
\textbf{Model} & \textbf{HAM10000} & \textbf{ISIC 2018} & \textbf{CR-AI4SkIN} \\
 & \textit{(ACC / F1-score)} & \textit{(ACC / F1-score)} & \textit{(ACC / F1-score)} \\
\hline
ResNet50    & 83.8\% / 84.0\% & 76.8\% / 77.0\% & 83.0\% / 84.0\% \\
MobileNet   & 83.1\% / 83.0\% & 73.2\% / 73.0\% & 80.0\% / 81.0\% \\
VGG16       & 82.9\% / 83.0\% & 75.3\% / 75.0\% & 76.0\% / 77.0\% \\
VGG19       & 81.5\% / 82.0\% & 74.9\% / 75.0\% & 72.0\% / 74.0\% \\
InceptionV3 & 80.1\% / 80.0\% & 70.8\% / 71.0\% & 78.0\% / 79.0\% \\
\hline
\end{tabular}
\caption{Accuracy and average F1-score for each model across the three datasets.}
\label{tab:acc-f1score}
\end{table}

\begin{table}[h!]
\centering
\begin{tabular}{lccc}
\hline
\textbf{Model} & \textbf{HAM10000} & \textbf{ISIC 2018} & \textbf{CR-AI4SkIN} \\
 & \textit{(Precision / Recall)} & \textit{(Precision / Recall)} & \textit{(Precision / Recall)} \\
\hline
ResNet50    & 84\% / 84\% & 77\% / 77\% & 84\% / 83\% \\
MobileNet   & 83\% / 83\% & 73\% / 73\% & 83\% / 80\% \\
VGG16       & 83\% / 83\% & 75\% / 75\% & 79\% / 76\% \\
VGG19       & 82\% / 82\% & 75\% / 75\% & 78\% / 72\% \\
InceptionV3 & 80\% / 80\% & 71\% / 71\% & 81\% / 78\% \\
\hline
\end{tabular}
\caption{Precision and Recall (Sensitivity) obtained by each model across the HAM10000, ISIC 2018, and CR-AI4SkIN datasets.}
\label{tab:precision-recall}
\end{table}

The results presented in \autoref{tab:acc-f1score} demonstrate that the ResNet50 architecture achieved the best performance measseures across all evaluated datasets. It reached an accuracy of 83.8\% and an macro F1-score of 84.0\% on HAM10000, 76.8\%/77.0\% on ISIC 2018, and 83.0\%/84.0\% on CR-AI4SkIN. This strong performance can be explained by the ResNet50 architecture, which employs residual connections that help capture image features more effectively, particularly in skin lesion analysis.

The MobileNet model also presented satisfactory results, ranking second in most experiments, with the exception of ISIC 2018, where VGG16 and VGG19 achieved higher accuracy. This finding is particularly relevant because MobileNet is a lightweight and computationally efficient model, indicating that competitive performance can be achieved even with less complex architectures.

It was observed that all models performed worse on the ISIC 2018 dataset, with results approximately 7\% to 10\% lower compared to the other datasets. This suggests that ISIC 2018 images may be more challenging, either due to greater variability in lesion appearance, image acquisition conditions, or class distribution differences.

A positive aspect observed is that all models maintained similar values of accuracy and macro F1-score, indicating a satisfactory balance between correctly identifying lesions (precision) and detecting the majority of relevant cases (recall), as also confirmed by the precision and recall values shown in Table \ref{tab:precision-recall}. This confirms that the approach based on feature extraction combined with supervised metric learning is effective for skin cancer classification.

From a more detailed perspective, the recall values indicate that most models were able to detect a large proportion of malignant lesions, which is a crucial aspect in medical diagnosis. However, errors still occur mainly in borderline cases where benign and malignant lesions share similar visual characteristics. These errors are more evident in architectures such as VGG16 and VGG19, which show a slightly larger drop in recall and macro F1-score, suggesting a lower capability to capture subtle discriminative patterns in more complex datasets.

The joint analysis of the metrics also highlights differences in the clinical robustness of the evaluated architectures. The balanced values of accuracy, precision, recall, and macro F1-score across all models are positive, as they contribute to reducing false negatives, which is a critical aspect in melanoma diagnosis. However, the performance drops observed mainly in the VGG architectures and in MobileNet under domain shifts indicate lower reliability of these models in more heterogeneous clinical scenarios. In contrast, ResNet50 demonstrates greater stability across different datasets, suggesting a more consistent clinical potential.

Beyond individual comparisons per dataset, the results also allow us to observe how the architectures behave when transitioning between imaging modalities. Models such as ResNet50 and VGG16 exhibit moderate variations between dermatoscopic and histopathological scenarios, indicating better cross-domain stability compared to the remaining architectures. In contrast, VGG19 and MobileNet show more pronounced performance drops across datasets, suggesting a reduced generalization capability when confronted with substantially different visual patterns. These findings reinforce that modality shifts have a direct impact on performance, affecting each architecture in a distinct manner.

After training and evaluating the models on each dataset, the Friedman test was applied to the accuracy metric to verify whether the performance differences among the architectures were statistically significant \citep{Friedman}. This non-parametric test is suitable for comparing multiple methods evaluated under the same experimental conditions, especially when the normality assumptions cannot be guaranteed, as is often the case in machine learning studies involving medical images.

As presented in Table~\ref{tab:friedman}, a statistically significant difference was found for all datasets, with p-values lower than 0.05. This indicates that the models do not exhibit equivalent performance when evaluated under the same experimental conditions. Both the ISIC 2018 and CR-AI4SkIN datasets present extremely low p-values ($<$0.0001), while HAM10000 also shows a significant result (p~=~0.0077). This indicates that the performance differences among architectures are statistically pronounced across all evaluated scenarios, being particularly strong in the more challenging datasets.

\begin{table}[h!]
\centering
\begin{tabular}{lc}
\hline
\textbf{Dataset} & \textbf{p-value} \\
\hline
HAM10000   & 0.0077 \\
ISIC 2018  & $<$0.0001 \\
CR-AI4SkIN & $<$0.0001 \\
\hline
\end{tabular}
\caption{Results of the Friedman test for accuracy across the three datasets.}
\label{tab:friedman}
\end{table}

Given the global differences indicated by the Friedman test, the Nemenyi post-hoc test was applied to identify which pairs of models exhibited statistically significant differences \citep{Nemenyi}. This procedure performs pairwise comparisons and considers two methods to differ significantly when the difference between their average ranks exceeds the critical difference, complementing the analysis by specifically indicating which architectures stand out relative to the others.

According to Table~\ref{tab:NemenyiHAM}, no pair of models reached statistical significance in the post-hoc analysis --- all p-values remain well above 0.05, with the lowest value being 0.487 for the ResNet50 $\times$ InceptionV3 comparison. Therefore, with 95\% confidence, no architecture can be considered statistically superior to the others on the HAM10000 dataset. Although the Friedman test detected a global difference, the Nemenyi results indicate that this difference is distributed diffusely across architectures, with no single model standing out as a clear outlier.

\begin{table}[h!]
\centering
\small
\begin{tabular}{lccccc}
\hline
\textbf{Model} & \textbf{ResNet50} & \textbf{MobileNet} & \textbf{VGG16} &
\textbf{VGG19} & \textbf{InceptionV3} \\
\hline
ResNet50    & 1.000000 & 0.997802 & 0.994397 & 0.863732 & 0.486649 \\
MobileNet   & 0.997802 & 1.000000 & 0.999987 & 0.964039 & 0.691827 \\
VGG16       & 0.994397 & 0.999987 & 1.000000 & 0.978412 & 0.744179 \\
VGG19       & 0.863732 & 0.964039 & 0.978412 & 1.000000 & 0.969382 \\
InceptionV3 & 0.486649 & 0.691827 & 0.744179 & 0.969382 & 1.000000 \\
\hline
\end{tabular}
\caption{Nemenyi post-hoc test --- p-values between models on the HAM10000 dataset.}
\label{tab:NemenyiHAM}
\end{table}

According to Table~\ref{tab:NemenyiISIC}, the only statistically significant difference found on the ISIC 2018 dataset is the comparison ResNet50 $\times$ InceptionV3 (p=0.017). Therefore, with 95\% confidence, we can confirm that ResNet50 and InceptionV3 exhibit distinct accuracy levels on this dataset. All other pairs yield p-values well above 0.05, indicating that MobileNet, VGG16, and VGG19 form a statistically homogeneous group together with each other and with both ResNet50 and InceptionV3 individually.

\begin{table}[h!]
\centering
\small
\begin{tabular}{lccccc}
\hline
\textbf{Model} & \textbf{ResNet50} & \textbf{MobileNet} & \textbf{VGG16} &
\textbf{VGG19} & \textbf{InceptionV3} \\
\hline
ResNet50    & 1.000000 & 0.755716 & 0.988802 & 0.969650 & 0.016941 \\
MobileNet   & 0.755716 & 1.000000 & 0.954733 & 0.980845 & 0.316869 \\
VGG16       & 0.988802 & 0.954733 & 1.000000 & 0.999889 & 0.068745 \\
VGG19       & 0.969650 & 0.980845 & 0.999889 & 1.000000 & 0.099962 \\
InceptionV3 & 0.016941 & 0.316869 & 0.068745 & 0.099962 & 1.000000 \\
\hline
\end{tabular}
\caption{Nemenyi post-hoc test --- p-values between models on the ISIC 2018 dataset.}
\label{tab:NemenyiISIC}
\end{table}

According to Table~\ref{tab:NemenyiCRAI}, the CR-AI4SkIN dataset presents the most pronounced statistical separations among all analyzed datasets. Statistically significant differences were found in all pairwise comparisons: ResNet50 $\times$ MobileNet (p$<$0.001), ResNet50 $\times$ VGG16 (p$<$0.001), ResNet50 $\times$ VGG19 (p$<$0.001), ResNet50 $\times$ InceptionV3 (p$<$0.001), MobileNet $\times$ VGG16 (p$<$0.001), MobileNet $\times$ VGG19 (p$<$0.001), MobileNet $\times$ InceptionV3 (p=0.010), VGG16 $\times$ VGG19 (p$<$0.001), VGG16 $\times$ InceptionV3 (p=0.016), and VGG19 $\times$ InceptionV3 (p$<$0.001). Therefore, with 95\% confidence, we can confirm that all pairs belong to distinct performance groups. This pattern demonstrates that on CR-AI4SkIN the architectures exhibit broadly distinct performance levels, with each architecture belonging to a distinct performance group.

\begin{table}[h!]
\centering
\small
\begin{tabular}{lccccc}
\hline
\textbf{Model} & \textbf{ResNet50} & \textbf{MobileNet} & \textbf{VGG16} &
\textbf{VGG19} & \textbf{InceptionV3} \\
\hline
ResNet50    & 1.000000 & 1.945e-04 & 1.110e-16 & 1.110e-16 & 5.770e-13 \\
MobileNet   & 1.945e-04 & 1.000000  & 2.014e-09 & 1.110e-16 & 1.020e-02 \\
VGG16       & 1.110e-16 & 2.014e-09 & 1.000000  & 7.482e-05 & 1.599e-02 \\
VGG19       & 1.110e-16 & 1.110e-16 & 7.482e-05 & 1.000000  & 3.290e-13 \\
InceptionV3 & 5.770e-13 & 1.020e-02 & 1.599e-02 & 3.290e-13 & 1.000000  \\
\hline
\end{tabular}
\caption{Nemenyi post-hoc test --- p-values between models on the CR-AI4SkIN dataset.}
\label{tab:NemenyiCRAI}
\end{table}

Regarding the results across datasets, the p-values confirm that all three datasets present statistically significant differences among the models. Both CR-AI4SkIN and ISIC~2018 yield substantially more significant results than HAM10000, which is consistent with the greater visual complexity introduced by different image acquisition conditions and dataset characteristics. While HAM10000 is composed of dermatoscopic images with relatively consistent visual and structural patterns, ISIC~2018 introduces higher variability in acquisition settings, and CR-AI4SkIN includes histopathological images with fundamentally different tissue organization. These differences lead the architectures to respond more divergently, producing more pronounced statistical contrasts.

\section{Conclusion}
\label{sec:Conclusion}

The results obtained allow us to conclude that ResNet50 consistently achieved the best performance across all three evaluated datasets, demonstrating greater stability across different image modalities. In the dermoscopic datasets HAM10000 and ISIC 2018, it is observed that all five architectures present statistically similar performances on HAM10000, forming a group of models with equivalent responses in this modality. On ISIC 2018, this homogeneity is partially maintained, with the only statistically significant distinction being observed between ResNet50 and InceptionV3 (p=0.017), while the remaining architectures formed an equivalent group. However, when analyzing the histopathological dataset CR-AI4SkIN, this pattern no longer holds. The architectures exhibit much more pronounced differences, with all pairwise comparisons reaching statistical significance, revealing distinct performance groups and indicating that the structural complexity of histopathological images impacts the response of each model more intensively. Thus, while ResNet50 can be identified as the most stable and consistently superior architecture, performance differences among models are strongly modulated by the characteristics of the dataset used, with some architectures showing greater sensitivity to modality shifts than others.

Our work contemplates, at least, three relevant contributions. The first contribution lies in the comprehensive comparison of architectures applied to different image modalities, which makes it possible to observe how each model responds to important structural variations between dermoscopic and histopathological datasets. This analysis shows that architectures exhibiting similar behavior in one domain do not necessarily maintain this pattern when evaluated in a different visual context, as evidenced by the transition from the statistically homogeneous behavior observed on HAM10000 to the fully differentiated performance groups found on CR-AI4SkIN. The second contribution refers to the understanding of the influence of modality shift on the generalization capability of convolutional neural networks. The performance differences observed across the datasets reveal that the structural complexity of histopathological images tends to accentuate contrasts that are not as evident in dermoscopic datasets, affecting each architecture in a distinct manner. The third contribution concerns the practical implications of these findings for computer-aided melanoma detection. The results demonstrate that conclusions drawn from a single dataset may not adequately reflect the real performance of models in diverse scenarios, reinforcing the importance of multi-dataset evaluations to identify architectures with greater stability and potential for use in computer-aided melanoma detection.

In summary, the results presented throughout this study indicate that the evaluation of models applied to melanoma detection requires an analysis that simultaneously considers different image modalities, as the performance achieved on one type of dataset does not guarantee equivalent behavior in other clinical contexts. The superior performance of ResNet50 across all evaluated scenarios, combined with its greater statistical separation from competing architectures particularly under domain shift, reinforces that model selection should consider not only isolated accuracy metrics but also stability, consistency, and adaptability across distinct datasets. In this sense, the findings contribute to expanding the understanding of the limitations and potential of the investigated architectures and highlight the importance of more comprehensive validation strategies capable of supporting the development of safer, more robust, and more reliable solutions for computer-aided melanoma diagnosis.

Future work may explore the integration of attention mechanisms and transformer-based architectures, which have demonstrated promising results in medical image analysis and may offer improved generalization across different imaging modalities. Additionally, investigating data augmentation strategies and domain adaptation techniques could help mitigate the performance drops observed under modality shifts, particularly for architectures such as VGG19 and MobileNet. The use of ensemble methods combining multiple architectures could also be investigated as a strategy to achieve more robust and consistent performance across dermoscopic and histopathological datasets. Finally, expanding the experimental evaluation to include additional datasets with greater class diversity and acquisition variability would contribute to a more comprehensive understanding of the generalization boundaries of the evaluated models in real-world clinical 
scenarios.

\section*{Acknowledgements}

The authors would like to thank the Coordination for the Improvement of Higher Education Personnel (CAPES) for the financial support.

\section*{CRediT authorship contribution statement}

\textbf{Wagner M. Schmitz}: conceptualization; formal analysis; methodology; software; writing - original draft; investigation. \textbf{Marco A. C. Barbosa}: writing – review \& editing; methodology. \textbf{Thiago M. Amaral}: writing – review \& editing; methodology. \textbf{D. Casanova}: writing – review \& editing; methodology; supervision. \textbf{J. T. Oliva}: writing – review \& editing; methodology; investigation; validation; supervision.

\section*{Declaration of generative AI use}

During the preparation of this manuscript, the authors used generative artificial intelligence tools exclusively to assist with the adaptation of author-drafted content into scientific language and to support the translation and synthesis of text originally developed by the authors. Generative AI was also used to assist in the search of related literature, supporting but not replacing the authors' own critical reading and interpretation of the reviewed works. At no point was generative AI used to independently generate content or draw conclusions.

\subsection*{Declaration of competing interest}
The authors declare that they have no competing financial interests or personal relationships that could have appeared to influence the work reported in this paper.

\section*{Data availability}
\phantomsection
\label{sec:dataavailability}

All datasets used in this study are publicly available. The HAM10000 dataset can be accessed at \url{https://dataverse.harvard.edu/dataset.xhtml?persistentId=doi:10.7910/DVN/DBW86T}; the ISIC 2018 Challenge dataset at \url{https://challenge.isic-archive.com/data/#2018}; and the CR-AI4SkIN dataset at \url{https://zenodo.org/records/10880652}.

\bibliographystyle{elsarticle-harv} 
\bibliography{cas-refs}

\end{document}